\pdfoutput=1

\documentclass[11pt]{article}

\usepackage[final]{acl}

\usepackage{times}
\usepackage{latexsym}
\usepackage{booktabs}
\usepackage[T1]{fontenc}

\usepackage{multirow}
\usepackage{graphicx}
\usepackage{framed}
\usepackage{dashrule}

\usepackage{arydshln}
\usepackage[table]{xcolor}
\usepackage[utf8]{inputenc}

\usepackage{stfloats}
\usepackage{microtype}
\usepackage{soul,xcolor}
\usepackage{inconsolata}
\usepackage[most]{tcolorbox}
\usepackage{graphicx}
\usepackage{marvosym}
\makeatletter
\newcommand\blfootnote[1]{
  \begingroup
  \renewcommand\thefootnote{}\footnote{#1}%
  \addtocounter{footnote}{-1}%
  \endgroup
}
\makeatother
\title{Quantifying the Impact of Structured Output Format on Large Language Models through Causal Inference}

\author{
Han Yuan\textsuperscript{†}, 
Yue Zhao\textsuperscript{†}, Li Zhang, Wuqiong Luo, Zheng Ma\textsuperscript{\Letter} \\
Global Decision Science, American Express  \\
\texttt{\{Han.Yuan1, Yue.Zhao, Li.Zhang1, Wuqiong.Luo, Zheng.Ma2\}@aexp.com}
}

\begin{document}
\maketitle
\begin{abstract}
Structured output from large language models (LLMs) has enhanced efficiency in processing generated information and is increasingly adopted in industrial applications. Prior studies have investigated the impact of structured output on LLMs' generation quality, often presenting one-way findings. Some suggest that structured format enhances completeness and factual accuracy, while others argue that it restricts the reasoning capacity of LLMs and leads to reductions in standard evaluation metrics. Potential limitations of these assessments include restricted testing scenarios, weakly controlled comparative settings, and reliance on coarse metrics. In this work, we present a refined analysis using causal inference. Based on one assumed and two guaranteed constraints, we derive five potential causal structures characterizing the influence of structured output on LLMs' generation: (1) collider without m-bias, (2) collider with m-bias, (3) single cause from instruction, (4) single cause from output format, and (5) independence. Across seven public and one developed reasoning tasks, we find that coarse metrics report positive, negative, or neutral effects of structured output on GPT-4o's generation. However, causal inference reveals no causal impact in 43 out of 48 scenarios. In the remaining 5, 3 involve multifaceted causal structures influenced by concrete instructions. Further experiments show that OpenAI-o3 are more resilient to output formats than general-purpose GPT-4o and GPT-4.1, highlighting an unaware advantage of reasoning models.
\end{abstract}
\blfootnote{\textsuperscript{†} These authors contributed equally to this work.} \blfootnote{\textsuperscript{\Letter} Correspondence: Zheng Ma, Singapore Decision Science Center of Excellence, American Express, 1 Marina Boulevard, 018989, Singapore.}

\section{Introduction}
Large language models (LLMs) have significantly improved accuracy and efficiency across various industrial applications. Their development has entered an era that encompasses not only foundation models but also complex systems integrating versatile modules. To explain how results are shaped by specific design choices, industrial applications demand impact analyses and clear documentation of newly introduced modules, option selections, and adjustments to existing components. Against this backdrop, we investigate how output formats affect the generation quality of LLMs and highlight that the proposed pipeline is not limited to structured outputs but can be extended to other modules.

\begin{figure}[t]
    \centering
\includegraphics[width=\columnwidth]{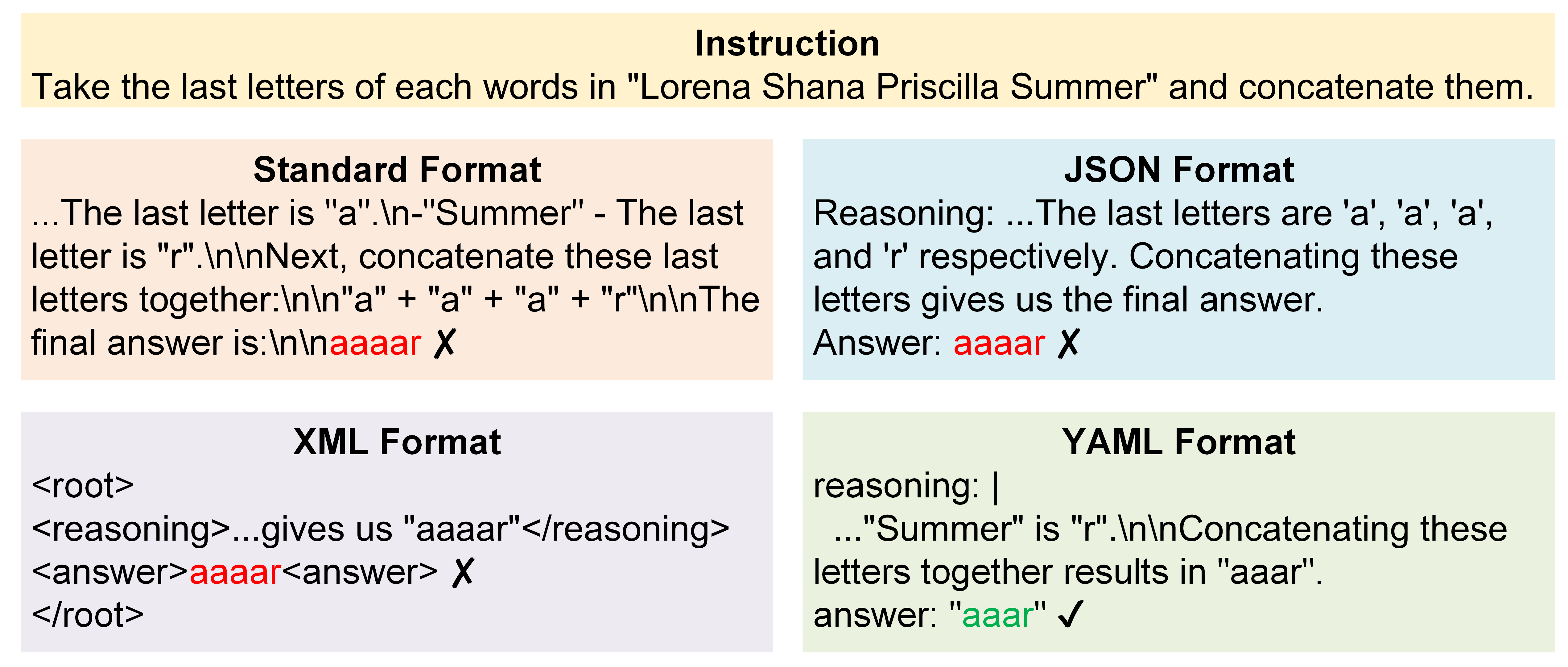}
    \caption{Positive, neutral, and negative effects of structured outputs on LLMs' generation.}
    \label{example}
\end{figure}


Structured output, defined as organizing generated content into a specific schema \citep{guo2025structuredoutputsenablegeneralpurpose,openai}, has been increasingly supported by LLMs, enabling them to directly output information in user-specified formats such as JSON, XML, and YAML\footnote{JSON: JavaScript Object Notation; XML: Extensible Markup Language; YAML: Yet Another Markup Language} \citep{tang-etal-2024-struc,wu-etal-2024-learning,guo-etal-2024-sample}. This development addresses a key limitation of earlier LLMs, which required users to either fine-tune models for structured generation \citep{chan-etal-2024-adapting,tang-etal-2024-struc} or apply post-processing techniques to convert unstructured outputs into the desired format \citep{chen-yang-2023-controllable,roegiest-etal-2023-questions,li-etal-2024-simple}. By eliminating this step, structured output enhances user experience and facilitates seamless engineering integration of LLMs into industrial applications such as Model Context Protocol (MCP) for agentic LLMs \citep{liu2024we,tam-etal-2024-speak,ayala-bechard-2024-reducing,yuan2025agentic}.

While structured output is widely adopted in industrial scenarios \citep{ayala-bechard-2024-reducing}, its impact on the generation quality of LLMs, compared to the standard unstructured format, remains insufficiently explored. Figure \ref{example} presents a real-world example from the Last Letter Concatenation task \citep{wei2022chain}, showing the inconsistent accuracies of different structured formats on the same case. Several prior studies have examined this problem but tend to reach one-sided conclusions that structured output either improves or impairs generation quality. \citet{guo2025structuredoutputsenablegeneralpurpose} identified two common challenges in medical question answering: unstructured outputs often contain factual inaccuracies and incomplete information, which are not fully resolved by the Chain of Thought (CoT) strategy \citep{wei2022chain}. To address this, they propose explicitly structuring the output generation process of LLMs, resulting in more factual and comprehensive responses. \citet{tam-etal-2024-speak} present a different perspective, reporting a decline in the reasoning performance of LLMs when generating structured outputs compared to free-form responses. Similar observations regarding the negative impact of structured output on generation quality have been reported in recent studies \citep{wang2025reassessing,jang2025instajudge,chen2025hierarchical,shi2025answering}. However, \citet{Will-Kurt} challenges the conclusions in \citet{tam-etal-2024-speak} by conducting experiments on the same datasets and deriving that structured generation improves LLMs' generation. This study attributes the contradictory findings to an experimental design in the earlier work: the failure to design comparable prompts for structured and unstructured formats, rendering the comparisons nonequivalent.

These studies focus on relatively narrow reasoning tasks and domains, leading to statements of uniformly positive or negative impacts that may not generalize. Also, some experiments lack direct comparability due to differences in prompt design, causing a loosely controlled evaluation across formats \citep{Will-Kurt}. In addition, these explorations apply similar strategy to investigate the impact of structured output that comparing the final aggregate metrics' differences between structured and unstructured output, which is relatively rudimentary and may obscure the nuanced and heterogeneous relationship \citep{causalai,hu2025extract}. Although an intuitive idea is to compute statistical differences between evaluation metrics under scenarios with and without structured formats, such methods remain at the bottom rung of causation ladder, and findings based solely on them lack sufficient rigor for application in highly regulated industries \citep{10.5555/3238230,jin2023cladder}. 

To address the limitations of previous studies: (1) the limited experimental scenarios, (2) incomparable prompts across output formats, and (3) incomprehensive impact assessments based on aggregate metrics' differences, this study presents a refined causal analysis of how structured output affects LLMs' generation across diverse reasoning tasks.

Specifically, we conduct experiments on seven widely used datasets to evaluate the effect of structured outputs on the diverse reasoning abilities of LLMs, including symbolic transformation, symbolic relations, factual verification, financial inference, system optimization and fault analysis, and code compilation. In addition, we introduce a new dataset, Enhanced Last Letter Concatenation (ELLC) based on the Last Letter Concatenation (LLC) dataset shown in Figure \ref{example}. LLC requires symbolic transformation of the final letters of given words, but with advances in LLMs, this benchmark has become insufficiently challenging. To address this limitation, ELLC incorporates a linguistic reasoning requirement of arranging the extracted letters into a valid English word. On this benchmark, GPT-4o achieves only 8.2\% accuracy in the most challenging sub-task, highlighting its difficulty. 

Moreover, we employ template conversion between formats, and leverage causal inference to reveal the impact of output formats, thereby ensuring a rigorous analysis. We categorize the impact into one of five candidate causal structures: (1) collider without m-bias, (2) collider with m-bias, (3) single cause from instruction, (4) single cause from output format, and (5) independence.

Our contributions are mainly fourfold: 
\begin{itemize}
    \item We reveal that structured output formats have no causal impact on GPT-4o's generation in the majority of reasoning scenarios;
    \item The remaining scenarios exhibit multifaceted effects of structured output formats, influenced by concrete users' instructions;
    \item We create a new dataset ELLC to challenge LLMs' capabilities in both symbolic transformation and linguistic reasoning;
    \item We present a causal analysis pipeline to evaluate a specific module's impact on LLMs, with applicability beyond structured outputs.
\end{itemize}

\begin{figure*}[t]
    \centering
\includegraphics[width=\textwidth]{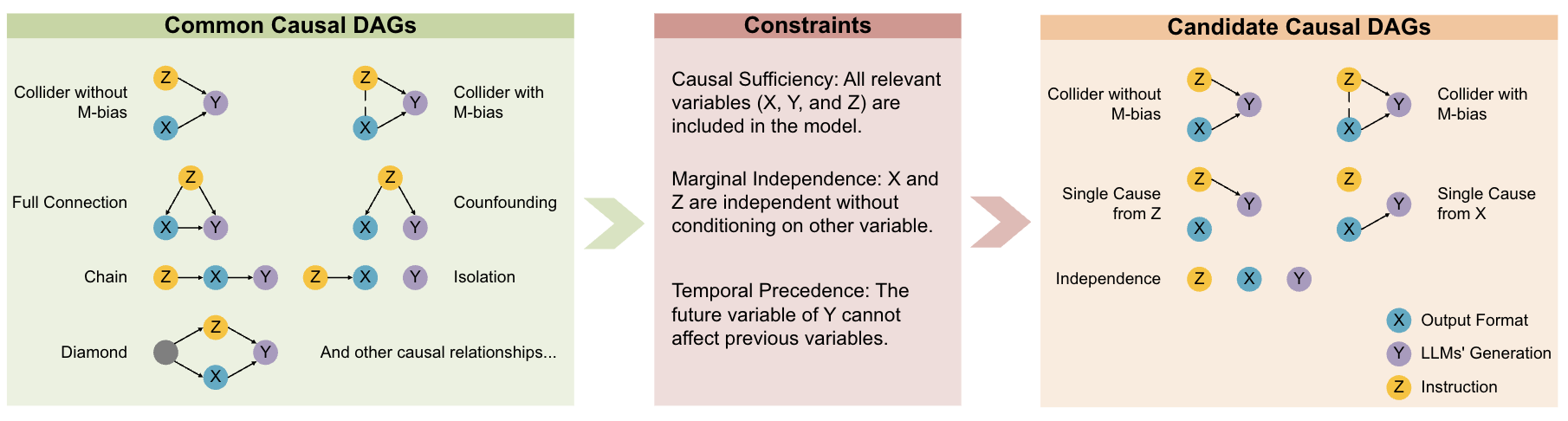}
    \caption{Candidate causal relationships extracted from common types based on constraints. For three variables of output format, LLMs' generation, and instruction, there could be diverse DAGs. Through detailed control and analysis, we satisfy the causal sufficiency and guarantee the other two, thereby reducing candidate DAGs to five.}
    \label{dags}
\end{figure*}

\section{Methodology}
Given the applications involving compliance and regulatory requirements in highly regulated industry, it is essential to conduct a detailed impact analysis. In our study, the adopted causal inference involves randomized controlled trials (RCTs), observational data collection, and statistical tests to determine the impact of structured output format on LLMs' generation \citep{10.5555/3238230}. In this section, we first present the variables involved in our analysis: instruction, output format, and LLMs' generation. The first two serve as potential causal factors, while the latter represents the outcome of interest. We then discuss the interventions targeting the two causal factors. Next, we elaborate on the potential causal structures using directed acyclic graphs (DAGs), which we aim to validate across various reasoning tasks. 
Finally, we outline the overall causal discovery. Our method is inspired by \citet{bao-etal-2025-likely} that explores CoT's impact on LLMs' generation through causal inference, with substantial modifications for our context. Practitioners can refer to our released analytical notebook \footnote{\url{https://github.com/Han-Yuan-Med/structured-output/blob/main/sot/sot_json.ipynb}} for detailed annotations and implementation procedures of the causal inference.

\subsection{Involved variables}
Following \citet{bao-etal-2025-likely}, we abstract the impact analysis into three primary variables: instruction, output format, and LLMs' generation, and control that other potentially influential factors. In addition to the three primary variables, the reasoning types represented in each dataset also influence the causal inference results and are naturally accounted for through the separate analysis of each dataset.

Instruction encompasses elements necessary to prompt LLMs' generation, including persona definition (system prompt), contextual information, problem formulation, and in-context learning cases \citep{swanson-etal-2021-story,miehling-etal-2025-evaluating}.

Output format is the outcome of interest. Prior studies present conflicting views on the effect of structured output format in LLMs' reasoning. Our study focuses on three structured formats of JSON, XML, and YAML, and compares them with the standard format of unstructured text.

LLMs' generation consists of two components: reasoning and answer. Per definitions by \citet{openai}, reasoning refers to the logical steps leading to the final conclusion, while answer denotes the final answer, taking into account the reasoning steps. We focus on answer as a proxy for LLMs' generation quality, leaving reasoning for future explorations.

\subsection{Interventions of instruction}
To manage experimental complexity, we follow \citet{bao-etal-2025-likely}, focus on zero-shot learning, do not incorporate external contexts such as knowledge graphs or web search tools, and limit alternative instruction in the main text to paraphrasing the system prompt into diverse persona definitions (Appendix \ref{prompts}), which is a practical and used method of instruction perturbation in latest studies \citep{hu2024quantifying,dong2024can,wu2025personas,li2025hello}. Additional interventions on context description and problem formulation, along with the results, are reported in Appendix \ref{additional_interventions}.

\subsection{Interventions of output format}
To enable LLMs to generate responses in specified formats of JSON, XML, or YAML, we adopt the same format-restricting instruction strategy \citep{tam-etal-2024-speak,zhao-etal-2024-longagent,zhao-etal-2024-longrag} by explicitly specifying the required structured output format in the model prompt. This approach ensures transparency and minimizes the introduction of noise into causal analysis. To ensure that instructions are comparable across the standard and structured formats, we use template conversion and the same components as \citet{openai} and \citet{tam-etal-2024-speak}, which include a pair of reasoning and answer (Figure \ref{template}). 
Although more complex structures, such as those with hierarchical architectures, can be adopted in the reasoning process, the primary focus of our study is to explore LLMs' generation with and without structured output, rather than with varying complexity.

In addition to the straightforward strategy, which is compatible with any LLM for obtaining structured output, we also assess JSON output via function calling, as detailed in the Appendix \ref{function_calling}. It is worth noting that, except for the experiments in Appendix \ref{function_calling}, all other experiments employ the format-restricting instruction strategy. Code for both format-restricting instruction and function calling settings is available on GitHub\footnote{\url{https://github.com/Han-Yuan-Med/structured-output/tree/main/generation}}.

\begin{figure}[t]
    \centering
\includegraphics[width=\columnwidth]{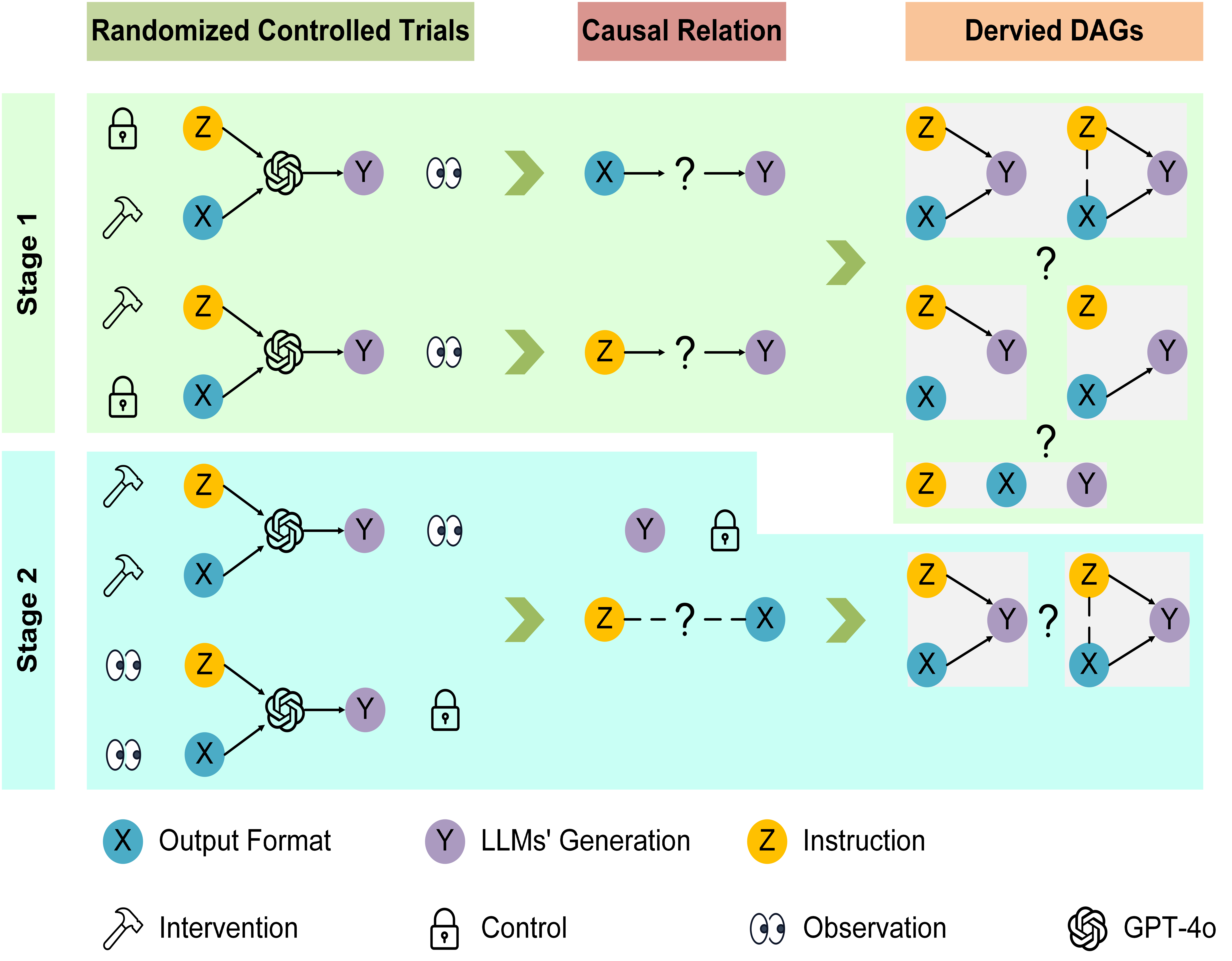}
    \caption{RCTs for identifying causal structures.}
    \label{rct}
\end{figure}

\subsection{Causal directed acyclic graphs}
\label{section_dags}
To depict causal structures,  we use causal directed acyclic graphs (DAGs); however, our types extend those explored in \citet{bao-etal-2025-likely} through more rigorous statistical designs.

Figure \ref{dags} shows our simplification process that reduces a large set of DAGs to a limited number of candidate structures, enabled by one controlled and two guaranteed constraints. The controlled constraint is causal sufficiency, which posits that DAGs are determined solely by the three variables illustrated above: instruction, output format, and LLMs' generation. This constraint is satisfied through our explicit control of other potentially influential factors within each scenario. The two guaranteed constraints are marginal independence and temporal precedence. In our setup, output format is  independent of instruction, as it remains constant across different tasks. Additionally, model generation occurs temporally after the specification of instruction and output format. Together, these properties allow us to avoid ambiguities introduced by the Markov Equivalence Class (MEC), where a given set of correlations 
can correspond to multiple causal relationships \citep{jin2024can}. Finally,  we derive 5 candidate causal DAGs, including collider without m-bias (CwoM), collider with m-bias (CwM), single cause from instruction (INS), single cause from output format (FMT), and independence (IND).

\subsection{Causal discovery}
\label{section_discovery}
To ensure quantitative causal inference, we employ a variety of statistical tools to uncover the structures depicted by the DAGs among involved variables and to assess the impact of structured output formats on LLMs' generation (Appendix \ref{statistical_test}). Generally, the discovery process involves two stages. The first stage determines whether output format $X$ or instruction $Z$ causally causes LLMs' generation by examining whether statistically significant differences exist in the target distributions with and without corresponding interventions $I$ from $X$ or $Z$: $E\left[Y\mid do(I)\right]-E\left[Y\right]$ \citep{rubin1974estimating,bao-etal-2025-likely}. If both factors have a significant impact, the second stage tests for an association between them conditioned on LLMs' generation. 

At the start, we investigate whether the structured output format exerts a statistically significant causal effect on LLMs' generation. The standard way \citep{bao-etal-2025-likely} is to control the instruction to the standard version, intervene on the output format using JSON, XML, and YAML, respectively, and compute the corresponding $p$-values using McNemar's test \citep{mcnemar1947note}, a reduced form of Cochran's Q test. We further refine our analysis by recognizing that instruction itself can be subject to intervention in our study. Treating the standard version as a single stratum risks overlooking heterogeneity across instruction variants. To mitigate this, we pool evidence across all instruction variants, ensuring our conclusions are not biased or limited to a single subset. There are two general approaches to address this: combining $p$-values from McNemar's test across each instruction stratum, or using the Cochran–Mantel–Haenszel test \citep{cochran1954some}. We employ the former approach because the latter one assumes independent samples across strata, which is not hold in our setting. For $p$-value combination, we use Stouffer's method \citep{stouffer1949american} due to its explicit coverage of correlated test statistics arising from shared samples across strata. The null hypothesis A states that there is no difference in LLMs' generation between the standard and a specific structured format across all strata. The alternative hypothesis asserts that a difference exists in at least one stratum. If the combined $p$-value is less than a specified significance level $\alpha$, we reject the null hypothesis and conclude that the specified format has a significant effect.

Similar to the output format, the causal effect of instruction on LLMs' generation is evaluated under null hypothesis B, stating that instruction does not influence LLMs' generation. To validate the effect, we fix the output format, intervene on instruction, apply statistical tests within each stratum, and aggregate the $p$-values for overall assessment. Different from the former binary intervention on the standard format with a specific schema, the instruction intervention includes four distinct classes, which necessitates Cochran's Q test \citep{cochran1950comparison}.

\begin{table*}[]
\resizebox{\textwidth}{!}{%
\begin{tabular}{cllcccccccc}
\hline
Hypotheses & Fixed modules & Examined modules & GSM8K & LLC & ELLC & SOT & GCF & GCC & OpsEval & XCodeEval \\ \hline
\multirow{4}{*}{A} & \multirow{4}{*}{\begin{tabular}[c]{@{}l@{}}Standard \& alternative\\ instruction\end{tabular}} & \begin{tabular}[c]{@{}l@{}}Controlled\\ (w/ standard format)\end{tabular} & 0.954 & 0.940 & 0.886 & 0.992 & 0.693 & 0.484 & 0.250 & 0.831 \\
 &  & \begin{tabular}[c]{@{}l@{}}Intervened\\ (w/ JSON format)\end{tabular} & \begin{tabular}[c]{@{}c@{}}0.951\\ (0.620)\end{tabular} & \begin{tabular}[c]{@{}c@{}}0.956\\ (0.387)\end{tabular} & \begin{tabular}[c]{@{}c@{}}0.875\\ (0.251)\end{tabular} & \begin{tabular}[c]{@{}c@{}}0.965\\ (0.030)\end{tabular} & \begin{tabular}[c]{@{}c@{}}0.700\\ (0.582)\end{tabular} & \begin{tabular}[c]{@{}c@{}}0.495\\ (0.250)\end{tabular} & \begin{tabular}[c]{@{}c@{}}0.330\\ (0.179)\end{tabular} & \begin{tabular}[c]{@{}c@{}}0.826\\ (0.779)\end{tabular} \\
 &  & \begin{tabular}[c]{@{}l@{}}Intervened\\ (w/ XML format)\end{tabular} & \begin{tabular}[c]{@{}c@{}}0.958\\ (0.711)\end{tabular} & \begin{tabular}[c]{@{}c@{}}0.952\\ (0.667)\end{tabular} & \begin{tabular}[c]{@{}c@{}}0.876\\ (0.300)\end{tabular} & \begin{tabular}[c]{@{}c@{}}0.974\\ (0.181)\end{tabular} & \begin{tabular}[c]{@{}c@{}}0.689\\ (0.224)\end{tabular} & \begin{tabular}[c]{@{}c@{}}0.490\\ (0.162)\end{tabular} & \begin{tabular}[c]{@{}c@{}}0.230\\ (0.883)\end{tabular} & \begin{tabular}[c]{@{}c@{}}0.818\\ (0.263)\end{tabular} \\
 &  & \begin{tabular}[c]{@{}l@{}}Intervened\\ (w/ YAML format)\end{tabular} & \begin{tabular}[c]{@{}c@{}}0.954\\ (0.512)\end{tabular} & \begin{tabular}[c]{@{}c@{}}0.977\\ (0.039)\end{tabular} & \begin{tabular}[c]{@{}c@{}}0.883\\ (0.344)\end{tabular} & \begin{tabular}[c]{@{}c@{}}0.989\\ (0.638)\end{tabular} & \begin{tabular}[c]{@{}c@{}}0.713\\ (0.287)\end{tabular} & \begin{tabular}[c]{@{}c@{}}0.487\\ (0.428)\end{tabular} & \begin{tabular}[c]{@{}c@{}}0.225\\ (0.903)\end{tabular} & \begin{tabular}[c]{@{}c@{}}0.791\\ (0.051)\end{tabular} \\ \hline
\multirow{6}{*}{B} & \multirow{2}{*}{Standard \& JSON format} & \begin{tabular}[c]{@{}l@{}}Controlled\\ (w/ standard instruction)\end{tabular} & 0.945 & 0.953 & 0.881 & 0.990 & 0.698 & 0.492 & 0.338 & 0.845 \\
 &  & \begin{tabular}[c]{@{}l@{}}Intervened\\ (w/ alternative instruction)\end{tabular} & \begin{tabular}[c]{@{}c@{}}0.954\\ (0.273)\end{tabular} & \begin{tabular}[c]{@{}c@{}}0.947\\ (0.159)\end{tabular} & \begin{tabular}[c]{@{}c@{}}0.880\\ (0.102)\end{tabular} & \begin{tabular}[c]{@{}c@{}}0.976\\ (0.019)\end{tabular} & \begin{tabular}[c]{@{}c@{}}0.696\\ (0.556)\end{tabular} & \begin{tabular}[c]{@{}c@{}}0.488\\ (0.713)\end{tabular} & \begin{tabular}[c]{@{}c@{}}0.278\\ (0.398)\end{tabular} & \begin{tabular}[c]{@{}c@{}}0.824\\ (0.379)\end{tabular} \\ \cline{2-11} 
 & \multirow{2}{*}{Standard \& XML format} & \begin{tabular}[c]{@{}l@{}}Controlled\\ (w/ standard instruction)\end{tabular} & 0.958 & 0.947 & 0.889 & 0.990 & 0.772 & 0.495 & 0.275 & 0.825 \\
 &  & \begin{tabular}[c]{@{}l@{}}Intervened\\ (w/ alternative instruction)\end{tabular} & \begin{tabular}[c]{@{}c@{}}0.956\\ (0.248)\end{tabular} & \begin{tabular}[c]{@{}c@{}}0.946\\ (0.117)\end{tabular} & \begin{tabular}[c]{@{}c@{}}0.879\\ (0.092)\end{tabular} & \begin{tabular}[c]{@{}c@{}}0.981\\ (0.095)\end{tabular} & \begin{tabular}[c]{@{}c@{}}0.683\\ (0.175)\end{tabular} & \begin{tabular}[c]{@{}c@{}}0.485\\ (0.023)\end{tabular} & \begin{tabular}[c]{@{}c@{}}0.231\\ (0.532)\end{tabular} & \begin{tabular}[c]{@{}c@{}}0.824\\ (0.645)\end{tabular} \\ \cline{2-11} 
 & \multirow{2}{*}{Standard \& YAML format} & \begin{tabular}[c]{@{}l@{}}Controlled\\ (w/ standard instruction)\end{tabular} & 0.953 & 0.953 & 0.886 & 0.988 & 0.701 & 0.493 & 0.250 & 0.812 \\
 &  & \begin{tabular}[c]{@{}l@{}}Intervened\\ (w/ alternative instruction)\end{tabular} & \begin{tabular}[c]{@{}c@{}}0.954\\ (0.143)\end{tabular} & \begin{tabular}[c]{@{}c@{}}0.960\\ (0.099)\end{tabular} & \begin{tabular}[c]{@{}c@{}}0.884\\ (0.296)\end{tabular} & \begin{tabular}[c]{@{}c@{}}0.991\\ (0.148)\end{tabular} & \begin{tabular}[c]{@{}c@{}}0.703\\ (0.287)\end{tabular} & \begin{tabular}[c]{@{}c@{}}0.483\\ (0.394)\end{tabular} & \begin{tabular}[c]{@{}c@{}}0.234\\ (0.706)\end{tabular} & \begin{tabular}[c]{@{}c@{}}0.811\\ (0.353)\end{tabular} \\ \hline
\multirow{3}{*}{C} & \multirow{3}{*}{LLMs' generation} & JSON format & - & - & - & (0.693) & - & - & - & - \\
 &  & XML format & - & - & - & - & - & - & - & - \\
 &  & YAML format & - & (0.895) & - & - & - & - & - & - \\ \hline
\multicolumn{2}{c}{\multirow{3}{*}{Derived DAGs}} & JSON format & \textcolor{cyan!80!black}{IND}, \textcolor{cyan!80!black}{IND} & \textcolor{cyan!80!black}{IND}, \textcolor{cyan!80!black}{IND} & \textcolor{cyan!80!black}{IND}, \textcolor{cyan!80!black}{IND} & \textcolor{orange!80!black}{CwoM}, \textcolor{orange!80!black}{CwoM} & \textcolor{cyan!80!black}{IND}, \textcolor{cyan!80!black}{IND} & \textcolor{cyan!80!black}{IND}, \textcolor{cyan!80!black}{IND} & \textcolor{cyan!80!black}{IND}, \textcolor{cyan!80!black}{IND} & \textcolor{cyan!80!black}{IND}, \textcolor{cyan!80!black}{IND} \\
\multicolumn{2}{c}{} & XML format & \textcolor{cyan!80!black}{IND}, \textcolor{cyan!80!black}{IND} & \textcolor{cyan!80!black}{IND}, \textcolor{cyan!80!black}{IND} & \textcolor{cyan!80!black}{IND}, \textcolor{violet}{INS} & \textcolor{cyan!80!black}{IND}, \textcolor{violet}{INS} & \textcolor{cyan!80!black}{IND}, \textcolor{cyan!80!black}{IND} & \textcolor{violet}{INS}, \textcolor{violet}{INS} & \textcolor{cyan!80!black}{IND}, \textcolor{cyan!80!black}{IND} & \textcolor{cyan!80!black}{IND}, \textcolor{cyan!80!black}{IND} \\
\multicolumn{2}{c}{} & YAML format & \textcolor{cyan!80!black}{IND}, \textcolor{cyan!80!black}{IND} & \textcolor{green!60!black}{FMT}, \textcolor{orange!80!black}{CwoM} & \textcolor{cyan!80!black}{IND}, \textcolor{cyan!80!black}{IND} & \textcolor{cyan!80!black}{IND}, \textcolor{cyan!80!black}{IND} & \textcolor{cyan!80!black}{IND}, \textcolor{cyan!80!black}{IND} & \textcolor{cyan!80!black}{IND}, \textcolor{cyan!80!black}{IND} & \textcolor{cyan!80!black}{IND}, \textcolor{cyan!80!black}{IND} & \textcolor{cyan!80!black}{IND}, \textcolor{green!60!black}{FMT} \\ \hline
\end{tabular}%
}
\caption{Discovery of GPT-4o's DAGs based on structured output by format-restricting instruction. Exact match scores are reported for GSM8K, LLC, ELLC, and SOT, while accuracy scores are reported for GCF, GCC, OpsEval, and XCodeEval. The $p$-values are reported in parentheses, with 0.000 indicating a value smaller than 0.0005 due to rounding to three decimal places. Each setting shows two DAGs based on $\alpha$ values of 0.05 (first) and 0.1 (second).}
\label{main_table}
\end{table*}

After that, causal structures between LLMs' generation and output format or instruction, as shown in Figure \ref{rct}, can distinguish the DAG among INS, FMT, and IND, provided that no more than one significant causal effect is present. If both output format and instruction significantly influence LLMs' generation, an additional test is required to distinguish between the CwoM and CwM by evaluating null hypothesis C that there is no conditional association within each stratum of LLMs' generation.
We employ a mixed-effects logistic regression model \cite{mclean1991unified,fisher1919xv} to test the association between the output format and instruction. In this model, the conditioning variable, LLMs' generation, is treated as a random effect, allowing us to account for both inter-stratum variation and intra-stratum correlation. If any of the instruction-level coefficients are found to be significantly different from zero, this indicates a statistically significant association between output format and instruction, conditioning on LLMs' generation. If a statistically significant association is detected, it indicates the presence of m-bias, supporting a CwM structure. Otherwise, the absence of such an association corresponds to a CwoM structure.

In summary, we extend the causal discovery by \citet{bao-etal-2025-likely} in three key aspects: (i) restoring analysis to the original multi-class setting rather than reducing it to binary, (ii) pooling evidence across strata to enhance statistical power, and (iii) examining potential m-bias from collider variables. Further details are provided in the Appendix~\ref{statistical_test}.

\section{Experiments}
\subsection{Datasets}
We evaluate LLMs' capabilities across different reasoning types using 7 datasets: Grade School Math 8K (GSM8K) \citep{cobbe2021training} for mathematics, LLC for symbolic transformation \citep{wei2022chain}, Shuffled Objects Tracking (SOT) for symbolic relation \citep{BigBench}, German Credit Factuality (GCF) \citep{yuan2025exploring} for factual verification, German Credit Causality (GCC) \citep{yuan2025exploring} for financial inference, Operations-oriented Evaluation (OpsEval) for system optimization and fault analysis \citep{opseval}, and Execution-based Code Evaluation (XCodeEval) for code compilation \citep{xcodeeval}. We also develop a new dataset, ELLC\footnote{\url{https://github.com/Han-Yuan-Med/structured-output/tree/main/ellc_dataset}}, which extends the original LLC task. ELLC requires LLMs to first perform symbolic transformation to select correct letters, followed by linguistic reasoning to infer the sequence of selected letters to form a valid word. Exemplificative cases, development steps, and benchmark results of ELLC are detailed in Appendix \ref{dataset_case} and \ref{ellc_variants}.

\subsection{Models}
We evaluate GPT-4o, GPT-4.1, OpenAI-o3, and several small language models (SLMs), including GPT-oss-20B \citep{gptoss}, Llama-3.1-8B \citep{llama}, Phi-3.5-mini \citep{phi}, and Gemma-2-9B \citep{gemma} in the impact analysis of structured output. The temperature is fixed at 0 to eliminate randomness, and all other parameters remain at their default values. Unless otherwise specified, experiments use GPT-4o as the backbone. Experiments using GPT-4.1, OpenAI o3, and SLMs are reported in Appendices \ref{appendix_41_o3} and \ref{open_source}. For reproducibility, we release the model outputs, analytical notebooks, and resulting quantitative outcomes on GitHub\footnote{\url{https://github.com/Han-Yuan-Med/structured-output}}. 

\subsection{Results}

Table~\ref{main_table} summarizes causal discovery across diverse reasoning types. Causal structures are classified based on the rejection of null hypotheses A (instruction effect) and B (format effect). If both are rejected, we further test hypothesis C for m-bias.
As we consider two significance levels $\alpha$ of 0.05 and 0.1, two DAGs are derived. The significance levels here apply to individual tests, while for DAGs' discovery, which involves two or three $p$-values, a Bonferroni correction can be applied to control for Type I errors (Appendix~\ref{bonferroni_correction}).

We illustrate the causal discovery process using representative examples, with the first four based on GPT-4o and the last based on Llama-3.1-8B:
\begin{itemize}
    \item \textcolor{cyan!80!black}{IND} (independence): Observed in OpsEval-JSON using GPT-4o. Both $p$-values for instruction (0.179) and format (0.398) exceed $\alpha$ of 0.1, suggesting no causal effect.
    \item \textcolor{green!60!black}{FMT} (single cause from output format): Seen in XCodeEval-YAML at $\alpha$ of 0.1. The $p$-value for format is 0.051, while that for instruction is 0.353, indicating only format has an effect.
    \item \textcolor{violet}{INS} (single cause from instruction): Found in GCC-XML, where the instruction $p$-value is 0.023 and the format $p$-value is 0.162, suggesting only instruction has a causal impact.
    \item \textcolor{orange!80!black}{CwoM} (collider without m-bias): At $\alpha$ of 0.1, both factors become significant for GPT-4o in the LLC-YAML setup. The m-bias test yields a $p$-value of 0.895, supporting CwoM.
    \item \textcolor{red!90!black}{CwM} (collider with m-bias): Detected in GSM8K-JSON using Llama-3.1-8B. Both instruction and format are significant at $\alpha$ of 0.1, and conditioning on generation reveals correlation between them, indicative of m-bias.
\end{itemize}

\textbf{GPT-4o demonstrates robust insensitivity to structured output formats}. In 43 out of 48 causal structures (DAGs), structured outputs have no statistically significant effect on generation. However, this robustness is not universal. In the remaining 5 DAGs, 3 are CwoM, where both instruction and output format causally influence the model's output. The experimental results highlight the necessity of the used multifaceted analysis that should not be simplified to a single-factor perspective or to coarse comparisons of aggregated metrics. \textbf{GPT-4.1 follows trends similar to GPT-4o}, performing robustly in most settings but still exhibiting vulnerability on the SOT task. By contrast, \textbf{OpenAI-o3 is consistently resilient to structured outputs}, showing an underappreciated strength of reasoning-intensive models to general-purpose models.

\textbf{Function calling is a promising approach for obtaining structured outputs}. Although the main text and most experiments focus on format-restricting instructions due to their generalizability across models and output formats, Appendix~\ref{function_calling} shows that independence remains the predominant DAGs in structured outputs by function calling, while the generation quality outperform that from format-restricting instructions in 5 of the 8 datasets.

\textbf{SLMs are more sensitive to structured output formats}. As shown in Appendix~\ref{open_source}, 7 out of 24 DAGs exhibit causal effects, compared to 5 out of 48 for GPT-4o. Also, different from GPT-4o's 100\% success rate across all datasets, the success rate of SLMs can drop below 50\% in certain reasoning tasks. The differences in causal discovery results between GPT-4o and SLMs stem from a combination of structured output generation and reasoning-based problem-solving capabilities, both of which are strongly influenced by model scale and architectural design. As evidenced by GPT-oss-20B, the largest model among all SLMs, it demonstrates greater robustness to interventions in both instruction and output format.


\textbf{Derived DAGs are robust under additional interventions and API calls}. Specifically, additional interventions target context and problem formulation, yielding the same DAGs as those presented in Table~\ref{main_table} (Appendix~\ref{additional_interventions}). However, when conditioning on formats, we find that GPT-4o is more robust to changes in persona than to changes in context or formulation. Furthermore, despite GPT-4o API non-determinism, multiple trials yield consistent DAGs (Appendix~\ref{multiple_trials}).
For SLMs, consistency is ensured by configuring the decoding mechanism to greedy search.

\begin{figure}[t]
    \centering
\includegraphics[width=\columnwidth]{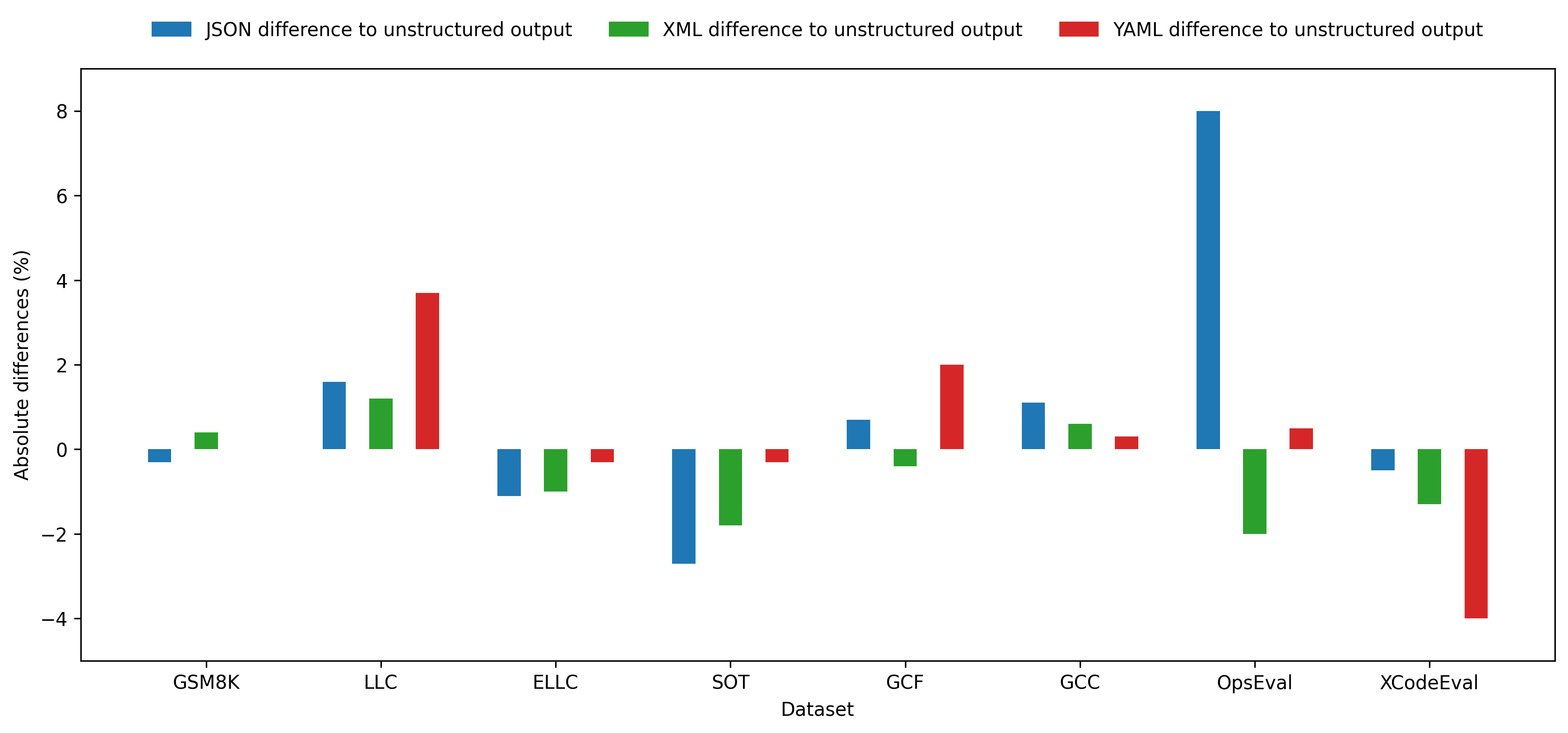}
    \caption{Comparison of evaluation metrics between unstructured and structured output formats.}
    \label{abs_rea}
\end{figure}

In summary, our causal inference pipeline provides a refined analysis on the impact of structured output formats. Relying solely on differences of evaluation metrics can be misleading, as the impact can be positive, negative, or neutral (Figure \ref{abs_rea}).
\section{Discussion and Conclusion}
Our study empirically reveals a multifaceted impact of structured output format on LLMs' reasoning performance, influenced by factors such as concrete instruction and reasoning type. These effects, as revealed by causal inference, are heterogeneous and nuanced, eluding prior coarse-grained approach solely focusing on aggregated evaluation metrics. Information from \citet{openai} helps illuminate potential reasons under the hood. To enable structured generation, GPT-4o employs three techniques: (1) a deterministic, engineering-driven approach that constrains its output, (2) fine-tuning the foundation model to comprehend complex schema and generate outputs that adhere to them, and (3) constrained decoding during generation that restricts sampling to tokens that conform to the specified schema rather than the full vocabulary. We hypothesize that the observed performance differences between structured and unstructured formats primarily arise from the latter two components: schema-aware fine-tuning and constrained decoding, which we envision that future development of foundation models can ensure generation quality under structured format matches that of standard free formats.

\section*{Limitations}
First, we acknowledge that some LLMs, such as Qwen \citep{yang2024qwen2} and DeepSeek \citep{liu2024deepseek}. Although we are unable to include these experiments, the introduced method is not constrained to a specific foundation model or module and is straightforward to apply to additional models seamlessly, offering a practical diagnostic tool for practitioners quantifying the impact of structured output formats.
Second, additional techniques, such as few-shot learning or reflection \citep{wang-etal-2024-math,zhang2024rest,gandhi2024stream,guan2025rstarmath,chen-etal-2025-graphcheck,qi2025mutual}, can enhance the performance of LLMs. Considering the complexity introduced by additional influential factors and the uncertainty caused by selected in-context cases or reflection mechanisms, we keep this direction as part of our future work. Third, we use answer accuracy as a proxy for generation quality. However, \citet{wang-etal-2024-math} have shown that LLMs can occasionally arrive at correct answers despite flawed reasoning. Future work will therefore examine the impact of structured output on LLMs' reasoning processes. Lastly, structured output can be achieved through other methods, such as output parsers, format-specific modes, etc. Our study focuses on format-restricting instructions and function calling because of their broad compatibility with most LLMs in industrial practice.

\section*{Ethical considerations}
The study offers an analytical framework for practitioners to understand the causal impact of structured output on LLMs' generation quality in their industrial applications. It highlights that structured output can affect LLMs' generation in certain reasoning scenarios. These insights may inspire future research in schema-aware fine-tuning and constrained decoding, aiming to achieve generation quality comparable to that of standard free formats.

\section*{Disclaimer}
This paper is provided solely for informational purposes as an academic contribution by the authors to the research community and does not represent, reflect, or constitute the views, policies, positions, or practices of American Express or its affiliates. Nothing in this paper should be cited or relied upon as evidence of, or support for, the business views, policies, positions, or practices of American Express or its affiliates.
\bibliography{custom}

\appendix
\section{Datasets and LLC development}
\label{dataset_case}
We conduct experiments on 8 distinct datasets to evaluate the impact of output format on LLMs' performance across diverse reasoning categories. Figure \ref{gsm8k_case}, \ref{llc_case}, \ref{ellc_case}, \ref{sot_case}, \ref{gcf_case}, \ref{gcc_case}, \ref{opseval_case}, and \ref{xcodeeval_case} present illustrative examples of Grade School Math 8K (GSM8K), Last Letter Concatenation (LLC), Enhanced Last Letter Concatenation (ELLC), Shuffled Objects Tracking (SOT), German Credit Factuality (GCF), and German Credit Causality (GCC),  Operations-oriented Evaluation (OpsEval), and Execution-based Code Evaluation (XCodeEval), respectively.

Specifically, GSM8K is used for mathematical reasoning, which contains 8,500 high-quality, linguistically diverse grade school math word problems authored by human experts. Each problem typically requires between two and eight steps to solve, primarily involving sequences of elementary arithmetic operations. For our experiments, we utilize the first 200 samples from the test set.

\begin{tcolorbox}[breakable, enhanced]
Question: Janet's ducks lay 16 eggs per day. She eats three for breakfast every morning and bakes muffins for her friends every day with four. She sells the remainder at the farmers' market daily for \$2 per fresh duck egg. How much in dollars does she make every day at the farmers' market?\\
\medskip
\noindent\dotfill\\
\medskip
Answer: Janet sells 16 - 3 - 4 = <<16-3-4=9>>9 duck eggs a day. She makes 9 * 2 = \$<<9*2=18>>18 every day at the farmer's market. \#\#\#\# 18
\end{tcolorbox}
\noindent\begin{minipage}{\columnwidth}
\captionof{figure}{An exemplificative case of GSM8K.}\label{gsm8k_case}
\end{minipage}

LLC is employed to evaluate LLMs' symbolic reasoning capabilities. The test set comprises 150 word-based puzzles, where the answer is generated by concatenating the last letters of specified words, a process that requires symbolic transformation. We also identified and corrected five labeling errors in the original dataset. Specifically, the label for "Claudia Cole Matthew Juan Pablo" was originally annotated as "aewo", but the correct label should be "aewno". For "Jorge Luis Mo Alexia Jerry", the label was "soay", whereas the correct answer is "esoay". In the case of "Miguel Angel Saul Brady Darryl", the original label "llyl" was missing a letter, and it has been corrected to "lllyl". Similarly, the label for "Charity Svetlana Jamie Jose A" was incorrectly given as "yaeA", which has now been revised to "yaeeA". Lastly, the label for "Marco Antonio Suzette Roland Isabel" was "oedl", but the accurate label is "ooedl". We use the revised test set with all 150 samples and corrected labels for our experiment and share it for readers' usage\footnote{\url{https://github.com/Han-Yuan-Med/structured-output/tree/main/llc/llc_test_revised.json}}.

\begin{tcolorbox}[breakable, enhanced]
Question: Take the last letters of each words in "Camilo Becky Eliza Rebecca" and concatenate them.\\
\medskip
\noindent\dotfill\\
\medskip
Answer: oyaa
\end{tcolorbox}
\noindent\begin{minipage}{\columnwidth}
\captionof{figure}{An exemplificative case of LLC.}\label{llc_case}
\end{minipage}

Building on LLC, we introduce a more complex task called ELLC. While LLC requires static concatenation of the last letters of given words, ELLC adds a linguistic reasoning component: models must determine the correct sequence of these letters to form a valid English word. We construct two versions: one using four input words, and another using six input words. Each sample includes a list of valid target words, as the same set of extracted letters can correspond to one or multiple correct answers. A response is considered correct if it matches any word in the target list. More challenging versions of this task involve a larger set of candidate words and letters, more complex extraction patterns, and the requirement for the model to generate all valid English words from the extracted letters. The main text reports experimental results for the basic task, while results for more complex variants are provided in the Appendix \ref{ellc_variants}.

We construct the ELLC dataset through the following steps. First, we extract unique items from the LLC dataset to form a basic pool, from which either the last or middle letters will be used for letter extraction. Next, we obtain common English words of four or six letters from the \textbf{NLTK Brown} corpus and map each letter in these words back to items in the initial pool. Since multiple items may correspond to a single letter, we sample one from the set of eligible items. We then identify a core set of combinations, where each combination consists of four or six items whose extracted letters collectively correspond to at least one valid English word. ELLC includes two letter extraction types: last letters, as in the original LLC dataset, and middle letters. For words with an odd number of letters, the middle letter is defined as the exact central character; for even-length words, it is defined as the left-central character. After that, for each combination, we enumerate all possible permutations of the selected letters and retain those that form valid English words, as verified against the broader \textbf{NLTK Words} corpus. This process yields a set of candidate items and a corresponding list of valid English words that can be formed from the specified letters. Further, we filter out samples whose target words appear only in NLTK Words but not in NLTK Brown, to exclude rare or obscure words and reduce ambiguity of ELLC task. Finally, we use Azure content management firewall to filter out samples that may contain words related to hate, self-harm, sex, or violence, ensuring responsible LLMs' generation that is not covered by NLTK.

\begin{tcolorbox}[breakable, enhanced]
Question: Take the last letters of each words in "Eunice Brayan Homero Herbert" and rearrange these selected letters to form valid English word(s), each word using each letter exactly once. The output may be a single word or multiple words, but must include every possible valid combination. What are the word(s)?\\
\medskip
\noindent\dotfill\\
\medskip
Answer: note, tone
\end{tcolorbox}
\noindent\begin{minipage}{\columnwidth}
\captionof{figure}{An exemplificative case of ELLC.}\label{ellc_case}
\end{minipage}

SOT evaluates a model's ability to infer the final state of a system given its initial configuration and a sequence of transformations. Each instance involves a set of objects initially assigned to individuals, followed by a series of pairwise exchanges. The model must then determine which object ends up with a given individual. The full dataset contains 3,750 samples spanning scenarios with three, five, and seven objects. For our experiments, we use the first 200 samples from the most challenging sub-task with seven objects.

\begin{tcolorbox}[breakable, enhanced]
Question: Alice, Bob, Claire, Dave, Eve, Fred, and Gertrude are playing a game. At the start of the game, they are each holding a ball: Alice has a green ball, Bob has a white ball, Claire has a yellow ball, Dave has a pink ball, Eve has a orange ball, Fred has a black ball, and Gertrude has a brown ball. As the game progresses, pairs of players trade balls. First, Bob and Gertrude swap balls. Then, Fred and Claire swap balls. Then, Dave and Gertrude swap balls. Then, Bob and Gertrude swap balls. Then, Alice and Claire swap balls. Then, Gertrude and Claire swap balls. Finally, Eve and Claire swap balls. At the end of the game, Alice has the\\
\medskip
\noindent\dotfill\\
\medskip
Answer: black ball
\end{tcolorbox}
\noindent\begin{minipage}{\columnwidth}
\captionof{figure}{An exemplificative case of SOT.}\label{sot_case}
\end{minipage}

GCF and GCC are two datasets designed to assess LLMs' ability to critically reflect on content generated by LLMs. This task is not easy, even for GPT-4o. A potential reason is that LLMs, despite being developed by different research institutions or companies, are often trained on overlapping corpora, such as Wikipedia. This overlap can result in high conditional probabilities being assigned to flawed outputs, making them appear more credible to LLMs than they do to humans \citep{gao-etal-2023-chatgpt}. Both datasets consist of the 1,782 reasoning points produced by LLMs in the context of financial decision-making, each annotated for the presence of factual or causal issues \citep{nli-as-a-judge}. 
All samples are included in our experiments.

\begin{tcolorbox}[breakable, enhanced]
Question: Does Checking account is low balance; Loan duration is longer than 50\% of all cases in the bank; Loan amount is higher than 50\% of all cases in the bank; Saving account is low balance; Present employment is shorter than 1 year; Installment rate in percentage of disposable income is lower than 50\% of all cases in the bank; The status of other debtors is none; Property is unknown or none; Age is older than 50\% of all cases in the bank; Other installment plans is none; Housing is not own; Job is a skilled employee or a resident; Number of people being liable to provide maintenance for is equal to or more than 1; Telephone is registered under the customer's name; Foreign worker is yes imply the customer's age is older than 50\% of the average bank customer, and they have a high number of people relying on them for maintenance?\\
\medskip
\noindent\dotfill\\
\medskip
Answer: No
\end{tcolorbox}
\noindent\begin{minipage}{\columnwidth}
\captionof{figure}{An exemplificative case of GCF.}\label{gcf_case}
\end{minipage}

\begin{tcolorbox}[breakable, enhanced]
Question: Does the customer has a low installment rate as a percentage of disposable income imply that the customer is bad credit?\\
\medskip
\noindent\dotfill\\
\medskip
Answer: No
\end{tcolorbox}
\noindent\begin{minipage}{\columnwidth}
\captionof{figure}{An exemplificative case of GCC.}\label{gcc_case}
\end{minipage}


OpsEval evaluates the operations and maintenance capabilities of LLMs in information technology systems, with a particular focus on industrial mobile communication networks. To ensure task difficulty, we exclude questions answerable through simple IT knowledge retrieval and instead include those requiring performance optimization or fault analysis. These tasks demand both technical understanding of system status and advanced reasoning to address practical industrial problems. For our experiments, we select all 40 questions that meet these criteria and share them on GitHub\footnote{\url{https://github.com/Han-Yuan-Med/structured-output/blob/main/opseval/mobile_communication_network_reasoning.json}}.

\begin{tcolorbox}[breakable, enhanced]
Question: When monitoring network traffic, you discover that the traffic for a certain period is abnormally high. How should you analyze and determine the cause?\\
A: Check the logs of all servers\\
B: Analyze the source and destination of the traffic data\\
C: Increase network bandwidth\\
D: Shut down unnecessary network services\\
\medskip
\noindent\dotfill\\
\medskip
Answer: B
\end{tcolorbox}
\noindent\begin{minipage}{\columnwidth}
\captionof{figure}{An exemplificative case of OpsEval.}\label{opseval_case}
\end{minipage}

XCodeEval \citep{xcodeeval} targets execution-based code understanding, generation, translation, and retrieval. We use the code compilation sub-task to assess models' comprehension of code content and their reasoning about whether the code can be successfully executed. Our evaluation includes 200 code examples across four programming languages, Kotlin, Go, Rust, and PHP, with each language represented by 25 compilable and 25 non-compilable cases to ensure balanced class distribution.

\begin{tcolorbox}[breakable, enhanced]
Question: Does the code generate a compilation error or not?\\
package "main"\\
import "fmt"\\
func main() \{\\
    \hspace*{1em}var n,k int\\
    \hspace*{1em}fmt.Scanf(\"\%d\%d", \&n, \&k)\\
    \hspace*{1em}for i := 0; i < n; i++ \{\\
        \hspace*{2em}if (5 * i * (i + 1) / 2 + k > 360) \{\\
            \hspace*{3em}fmt.Println(i)\\
            \hspace*{3em}return\\
        \hspace*{2em}\}\\
    \hspace*{1em}\}\\
    \hspace*{1em}fmt.Println(n)\\
\}\\
\medskip
\noindent\dotfill\\
\medskip
Answer: True
\end{tcolorbox}
\noindent\begin{minipage}{\columnwidth}
\captionof{figure}{An exemplificative case of XCodeEval.}\label{xcodeeval_case}
\end{minipage}

\section{Prompts}
\label{prompts}

Following \citet{tam-etal-2024-speak} and \citet{bao-etal-2025-likely}, we design both standard and intervened prompts, covering system persona definitions (Figure \ref{role}) and output formats (Figure \ref{template}). For contextual information and problem formulation, we retain the original content provided in the respective datasets. We focus on the zero-shot learning setting and therefore do not incorporate illustrative demonstrations of in-context learning examples. Specifically, for instruction interventions, we follow \citet{bao-etal-2025-likely} by paraphrasing the system prompt using different personas (i.e., professional roles). The standard role and task-specific parameters vary by dataset. For instance, in the GSM8K dataset, the standard role is a math teacher, and the task-specific problem refers to math problems. For output format interventions, we follow adopt two methods: format-restricting strategy as \citet{tam-etal-2024-speak} and function calling in Appendix \ref{function_calling}.

\begin{tcolorbox}[breakable, enhanced]
Standard\\
Please act as a \{standard role\} and solve the \{task specific\} problem step by step.\\
\medskip
\noindent\dotfill\\
\medskip
Chef\\
Imagine you are a chef in a bustling kitchen,
and you need to tackle the \{task specific\} problem as if
it were a recipe. Break down the solution into
clear, step-by-step instructions.
\medskip\\
\noindent\dotfill\\
\medskip
Detective\\
Imagine you are a detective unraveling a mystery. Solve the \{task specific\} problem meticulously, step by step, as you would piece together clues in an investigation.\\
\medskip
\noindent\dotfill\\
Judge\\
I need you to take on the role of a judge and adjudicate the \{task specific\} problem, providing a detailed step-by-step resolution.\\
\medskip
\noindent\dotfill\\
Artist\\
Imagine you are an artist, and approach solving the \{task specific\} problem with creativity and flair, breaking it down into steps.
\end{tcolorbox}
\noindent\begin{minipage}{\columnwidth}
\captionof{figure}{Interventions on instructions for GSM8K.}\label{role}
\end{minipage}

\begin{tcolorbox}[breakable, enhanced]
Unstructured prompt\\
Output the reasoning steps leading to the final conclusion and the final answer, taking into account the reasoning steps.\\
\medskip
\noindent\dotfill\\
\medskip
JSON format prompt\\
\{\\
"reasoning": "The reasoning steps leading to the final conclusion.",\\
"answer": "The final answer, taking into account the reasoning steps."\\
\}\\
\medskip
\noindent\dotfill\\
\medskip
XML format prompt\\
<root>\\
<reasoning>[The reasoning steps leading to the final conclusion.]</reasoning>\\
<answer>[The final answer, taking into account the reasoning steps.]<answer>\\
</root>\\
\medskip
\noindent\dotfill\\
YAML format prompt\\
\{\\
reasoning: |\\
\hspace*{1em}"<The reasoning steps leading to the final conclusion.>"\\
answer: "<The final answer, taking into account the reasoning steps.>"\\
\}
\end{tcolorbox}
\noindent\begin{minipage}{\columnwidth}
\captionof{figure}{Unstructured and structured format prompts.}\label{template}
\end{minipage}

\section{Statistical test}
\label{statistical_test}
Building on \citet{bao-etal-2025-likely}, we extend the causal discovery framework with more rigorous statistical considerations. This enhanced framework can serve as a reference for researchers interested in analyzing the nuanced effects of specific LLM modules on generation quality.

The first improvement we introduce is refining their simplification of a multi-class variable into a binary one, by restoring the analysis to the original multi-class setting. Specifically, they simplify a multi-class intervention $V^a=\{V_{1}^a,...,V_{M}^a\}$ into a binary one by randomly selecting one class per sample. While this approach reduces experimental costs, it risks obscuring meaningful effects: if one class positively influences outcomes while another does not, their opposing effects may cancel out when averaged, potentially leading to the false conclusion that the treatment has no overall impact. To address this issue, we conduct multi-arm RCTs and apply Cochran's Q test \citep{cochran1950comparison} to examine the statistical relationship between model performance and instruction or output format under diverse intervention conditions. The test's assumptions, including large sample approximation, random sample selection, and binary outcomes, are satisfied. The null hypothesis posits that different interventions have no effect on outcomes, while the alternative hypothesis asserts that differences exist among the interventions. A $p$-value below the specified threshold $\alpha$ leads to rejection of the null hypothesis, indicating that the intervened module has a causal relationship with LLMs' generation. It is worth noting that when only two interventions are compared, such as JSON format versus the standard unstructured format, Cochran's Q test reduces to McNemar's test \citep{mcnemar1947note}. 

\begin{table*}[]
\centering
\small
\begin{tabular}{lccccccccc}
\hline
Examined modules & Rho & GSM8K & LLC & ELLC & SOT & GCF & GCC & OpsEval & XCodeEval \\ \hline
\multirow{5}{*}{\begin{tabular}[c]{@{}l@{}}Intervened\\ (w/ JSON format)\end{tabular}} & 0.1 & 0.534 & 0.278 & 0.150 & 0.007 & 0.490 & 0.149 & 0.059 & 0.725 \\
 & 0.2 & 0.583 & 0.339 & 0.204 & 0.017 & 0.542 & 0.203 & 0.096 & 0.756 \\
 & 0.3 & 0.620 & 0.387 & 0.251 & 0.030 & 0.582 & 0.250 & 0.132 & 0.779 \\
 & 0.4 & 0.648 & 0.426 & 0.291 & 0.046 & 0.612 & 0.290 & 0.166 & 0.796 \\
 & 0.5 & 0.671 & 0.459 & 0.325 & 0.063 & 0.637 & 0.324 & 0.197 & 0.810 \\ \hline
\multirow{5}{*}{\begin{tabular}[c]{@{}l@{}}Intervened\\ (w/ alternative instruction)\end{tabular}} & 0.1 & 0.233 & 0.126 & 0.075 & 0.011 & 0.522 & 0.689 & 0.390 & 0.339 \\
 & 0.2 & 0.254 & 0.142 & 0.088 & 0.015 & 0.540 & 0.701 & 0.410 & 0.360 \\
 & 0.3 & 0.273 & 0.159 & 0.102 & 0.019 & 0.556 & 0.713 & 0.429 & 0.379 \\
 & 0.4 & 0.291 & 0.175 & 0.115 & 0.024 & 0.571 & 0.723 & 0.446 & 0.396 \\
 & 0.5 & 0.307 & 0.190 & 0.128 & 0.029 & 0.584 & 0.732 & 0.461 & 0.413 \\ \hline
\end{tabular}%
\caption{Sensitivity analysis of $\rho$ values on $p$-values in the discovery of GPT-4o's DAGs based on format-restricting instruction. A $p$-value of 0.000 indicates a value smaller than 0.0005, resulting from numerical rounding.}
\label{rho_sensitivity}
\end{table*}

Second, \citet{bao-etal-2025-likely} investigate the impact of a variable independently within each control setting, a complementary approach is to pool evidence across strata, which is known to increase statistical power and yield more precise estimates. Specifically, we implement statistical $p$-value combination across different strata of control scenarios.  A classic method is the Cochran–Mantel–Haenszel test \citep{cochran1954some}. However, it assumes independence of samples across strata, which does not hold in our study, as the same data instances are subjected to different interventions to form the strata. Therefore,  we opt to combine $p$-values from statistical tests conducted within each stratum to support aggregated inference. Without loss of generality, we consider correlation between multi-class variables $V^a=\{V_{1}^a,...,V_{M}^a\}$ and $V^b=\{V_{1}^b,...,V_{N}^b\}$, under the stratification of a  conditioning variable $V^c=\{V_{1}^c,...,V_{K}^c\}$. The data are first stratified by the levels of $V^c$, and within each stratum, a statistical test is conducted between 
$V^a$ and $V^b$, deriving a series of $p$-values $P_{1},...P_{k}$. Finally, $p$-values are aggregated for the overall association based on 
Stouffer's method \citep{stouffer1949american}. The null hypothesis posits no association within each stratum, while the alternative hypothesis suggests an association exists in at least one stratum. Stouffer's method holds an advantage over classic Fisher's method in that it can explicitly account for correlated test statistics arising from shared samples across strata. Such correlations can be empirically estimated by generating multiple observations per sample within each stratum, which achieved by setting the decoding temperature above zero and repeating the experiments. Alternatively, a practical approach is to adopt a rule-of-thumb correlation $\rho$ based on the knowledge that the strata in our setting are constructed from the same group of samples subjected to different interventions. The main text reports results based on a $\rho$ of 0.3, and Table~\ref{rho_sensitivity} provides additional statistical analyses using $\rho$ values of 0.1, 0.2, 0.4, and 0.5. Across 64 settings (8 datasets $\times$ 4 $\rho$ values $\times$ 2 statistical tests $\times$ 1 structured format of JSON), the conclusions remain consistent in 63 settings at an $\alpha$-value of 0.05 and in 60 settings at an $\alpha$-value of 0.1.

\begin{table*}[t]
\resizebox{\textwidth}{!}{%
\begin{tabular}{cllcccccccc}
\hline
Hypotheses & Fixed modules & Examined modules & GSM8K & LLC & ELLC & SOT & GCF & GCC & OpsEval & XCodeEval \\ \hline
\multirow{2}{*}{A} & \multirow{2}{*}{\begin{tabular}[c]{@{}l@{}}Standard \& alternative\\ instruction\end{tabular}} & \begin{tabular}[c]{@{}l@{}}Controlled\\ (w/ standard format)\end{tabular} & 0.954 & 0.940 & 0.886 & 0.992 & 0.693 & 0.484 & 0.250 & 0.831 \\
 &  & \begin{tabular}[c]{@{}l@{}}Intervened\\ (w/ JSON format)\end{tabular} & \begin{tabular}[c]{@{}c@{}}0.960\\ (0.415)\end{tabular} & \begin{tabular}[c]{@{}c@{}}0.971\\ (0.046)\end{tabular} & \begin{tabular}[c]{@{}c@{}}0.884\\ (0.630)\end{tabular} & \begin{tabular}[c]{@{}c@{}}0.978\\ (0.225)\end{tabular} & \begin{tabular}[c]{@{}c@{}}0.701\\ (0.542)\end{tabular} & \begin{tabular}[c]{@{}c@{}}0.494\\ (0.213)\end{tabular} & \begin{tabular}[c]{@{}c@{}}0.285\\ (0.697)\end{tabular} & \begin{tabular}[c]{@{}c@{}}0.823\\ (0.205)\end{tabular} \\ \hline
\multirow{2}{*}{B} & \multirow{2}{*}{Standard \& JSON format} & \begin{tabular}[c]{@{}l@{}}Controlled\\ (w/ standard instruction)\end{tabular} & 0.955 & 0.957 & 0.886 & 0.990 & 0.701 & 0.496 & 0.287 & 0.848 \\
 &  & \begin{tabular}[c]{@{}l@{}}Intervened\\ (w/ alternative instruction)\end{tabular} & \begin{tabular}[c]{@{}c@{}}0.958\\ (0.489)\end{tabular} & \begin{tabular}[c]{@{}c@{}}0.955\\ (0.249)\end{tabular} & \begin{tabular}[c]{@{}c@{}}0.885\\ (0.087)\end{tabular} & \begin{tabular}[c]{@{}c@{}}0.984\\ (0.359)\end{tabular} & \begin{tabular}[c]{@{}c@{}}0.696\\ (0.524)\end{tabular} & \begin{tabular}[c]{@{}c@{}}0.487\\ (0.525)\end{tabular} & \begin{tabular}[c]{@{}c@{}}0.263\\ (0.825)\end{tabular} & \begin{tabular}[c]{@{}c@{}}0.822\\ (0.152)\end{tabular} \\ \hline
C & LLMs' generation & JSON format & - & - & - & - & - & - & - & - \\ \hline
\multicolumn{2}{c}{Derived DAGs} & JSON format & \textcolor{cyan!80!black}{IND}, \textcolor{cyan!80!black}{IND} & \textcolor{green!60!black}{FMT}, \textcolor{green!60!black}{FMT} & \textcolor{cyan!80!black}{IND}, \textcolor{violet}{INS} & \textcolor{cyan!80!black}{IND}, \textcolor{cyan!80!black}{IND} & \textcolor{cyan!80!black}{IND}, \textcolor{cyan!80!black}{IND} & \textcolor{cyan!80!black}{IND}, \textcolor{cyan!80!black}{IND} & \textcolor{cyan!80!black}{IND}, \textcolor{cyan!80!black}{IND} & \textcolor{cyan!80!black}{IND}, \textcolor{cyan!80!black}{IND} \\ \hline
\end{tabular}%
}
\caption{Discovery of GPT-4o's DAGs based on structured output by function calling. The $p$-values are reported in parentheses. Each setting shows two DAGs based on significance levels of 0.05 (first) and 0.1 (second).}
\label{function_call_table}
\end{table*}

The third extension we introduce beyond prior work is the investigation of potential m-bias, a spurious association that arises when conditioning on a collider variable. Identifying the presence of m-bias allows us to distinguish between two causal structures: collider with m-bias (CwM) and collider without m-bias (CwoM). An intuitive approach would be to use Cochran's Q test with $p$-value combination, treating LLMs' generation quality as the stratification criterion, to assess whether the proportions of output formats across instruction variants remain consistent. However, this differs from earlier settings in two critical ways. First, samples within a stratum can be dependent: the same input may yield correct responses under both standard and structured formats. Second, Cochran's Q test is designed to evaluate the equality of intervention effects across conditions, rather than to test for associations between two categorical variables. These goals are conceptually and statistically distinct. Moreover, due to sample dependence, we cannot apply conventional statistical tests that assume independent observations, such as the Chi-square test \citep{pearson1900x}. To address this, we formulate the test of spurious association using a mixed-effects multinomial logistic regression to examine the association between instruction $V^a$ and output format $V^b$, conditioned on the outcome of LLMs' generation $V^c$. In our case, since $V^b$ is binary, the multinomial setting reduces to a binary logistic regression. Mathematically, we have the condition $V^c=\{V_{1}^c,...,V_{K}^c\}$, for each condition, we introduce a random intercept $I_k\sim N(0, \sigma^2)$ and model the association between instruction $V^a$ and output format $V^b$ conditioned on $V_{k}^c$:
\[
\text{logit}\,\Pr(V_{k}^b = 1) = \beta_0 + \sum_{m=1}^M \beta_m V_{m,k}^a + I_k.
\]
After the model fitting, we obtain the instruction-level coefficients $\hat{\beta}_m$ and compute Wald-type statistics $z_m$ as
$\hat{\beta}_m/{\text{SD}(\hat{\beta}_m)}$ with corresponding two-sided $p$-values as $2\left(1 - \Phi(|z_k|)\right)$, where $\Phi$ represents the cumulative distribution function. If any $p$-values are statistical significant, we identify a conditional association between output format and instruction, thus supporting the presence of m-bias and the CwM DAG. Otherwise, in the absence of such association, the causal structure is CwoM.

\section{Bonferroni correction}
\label{bonferroni_correction}
The causal discovery process in this paper relies on statistical methods to test the impacts of structured output and instruction, respectively. If both effects are significant, an additional test is conducted to identify potential m-bias. The significance levels of $\alpha = 0.05$ and $\alpha = 0.1$ reported in Table~\ref{main_table} apply to individual tests. However, for DAGs' discovery, which is based on two or three $p$-values, the significance levels should be adjusted to account for the increased risk of Type I errors (false positives) that arises when multiple hypotheses are used jointly to derive the final DAGs. A straightforward adjustment is the Bonferroni correction, which lowers the significance threshold by dividing the overall desired $\alpha$ level of 0.1 by the number of statistical tests. In our context, for a specific dataset and structured format, the denominator is 2 or 3 depending on the number of tests to derive the DAG.

If we were to draw a single overarching conclusion regarding whether structured output influences GPT-4o's generation performance by treating all scenarios in Table~\ref{main_table} as one study, the Bonferroni correction should be applied with a denominator of 50. However, we recommend conducting individual analyses across the 24 scenarios (8 datasets $\times$ 3 structured formats) for the following reasons. First, the six datasets differ in focus, as each evaluates distinct reasoning capabilities of LLMs. Second, the 3 structured output formats should not be aggregated, as separate analyses provide finer granularity and facilitate format-specific recommendations. This is particularly important when aiming to minimize the risk of replacing the default unstructured output with structured output in industrial applications.

\begin{table*}[b]
\resizebox{\textwidth}{!}{%
\begin{tabular}{cllccc}
\hline
Hypotheses & Fixed modules & Examined modules & SOT & OpsEval & XCodeEval \\ \hline
\multirow{4}{*}{A} & \multirow{4}{*}{\begin{tabular}[c]{@{}l@{}}Standard \& alternative\\ instruction\end{tabular}} & \begin{tabular}[c]{@{}l@{}}Controlled\\ (w/ standard format)\end{tabular} & 0.985 & 0.280 & 0.845 \\
 &  & \begin{tabular}[c]{@{}l@{}}Intervened\\ (w/ JSON format)\end{tabular} & \begin{tabular}[c]{@{}c@{}}0.859\\ (0.000)\end{tabular} & \begin{tabular}[c]{@{}c@{}}0.265\\ (0.839)\end{tabular} & \begin{tabular}[c]{@{}c@{}}0.842\\ (0.620)\end{tabular} \\
 &  & \begin{tabular}[c]{@{}l@{}}Intervened\\ (w/ XML format)\end{tabular} & \begin{tabular}[c]{@{}c@{}}0.961\\ (0.090)\end{tabular} & \begin{tabular}[c]{@{}c@{}}0.265\\ (0.684)\end{tabular} & \begin{tabular}[c]{@{}c@{}}0.855\\ (0.440)\end{tabular} \\
 &  & \begin{tabular}[c]{@{}l@{}}Intervened\\ (w/ YAML format)\end{tabular} & \begin{tabular}[c]{@{}c@{}}0.934\\ (0.001)\end{tabular} & \begin{tabular}[c]{@{}c@{}}0.290\\ (1.000)\end{tabular} & \begin{tabular}[c]{@{}c@{}}0.849\\ (0.692)\end{tabular} \\ \hline
\multirow{6}{*}{B} & \multirow{2}{*}{Standard \& JSON format} & \begin{tabular}[c]{@{}l@{}}Controlled\\ (w/ standard instruction)\end{tabular} & 0.945 & 0.300 & 0.848 \\
 &  & \begin{tabular}[c]{@{}l@{}}Intervened\\ (w/ alternative instruction)\end{tabular} & \begin{tabular}[c]{@{}c@{}}0.916\\ (0.000)\end{tabular} & \begin{tabular}[c]{@{}c@{}}0.266\\ (0.347)\end{tabular} & \begin{tabular}[c]{@{}c@{}}0.843\\ (0.751)\end{tabular} \\ \cline{2-6} 
 & \multirow{2}{*}{Standard \& XML format} & \begin{tabular}[c]{@{}l@{}}Controlled\\ (w/ standard instruction)\end{tabular} & 0.985 & 0.275 & 0.850 \\
 &  & \begin{tabular}[c]{@{}l@{}}Intervened\\ (w/ alternative instruction)\end{tabular} & \begin{tabular}[c]{@{}c@{}}0.970\\ (0.001)\end{tabular} & \begin{tabular}[c]{@{}c@{}}0.272\\ (0.495)\end{tabular} & \begin{tabular}[c]{@{}c@{}}0.850\\ (0.731)\end{tabular} \\ \cline{2-6} 
 & \multirow{2}{*}{Standard \& YAML format} & \begin{tabular}[c]{@{}l@{}}Controlled\\ (w/ standard instruction)\end{tabular} & 0.975 & 0.300 & 0.848 \\
 &  & \begin{tabular}[c]{@{}l@{}}Intervened\\ (w/ alternative instruction)\end{tabular} & \begin{tabular}[c]{@{}c@{}}0.956\\ (0.000)\end{tabular} & \begin{tabular}[c]{@{}c@{}}0.281\\ (0.533)\end{tabular} & \begin{tabular}[c]{@{}c@{}}0.847\\ (0.584)\end{tabular} \\ \hline
\multirow{3}{*}{C} & \multirow{3}{*}{LLMs' generation} & JSON format & (0.256) & - & - \\
 &  & XML format & (0.710) & - & - \\
 &  & YAML format & (0.435) & - & - \\ \hline
\multicolumn{2}{c}{\multirow{3}{*}{Derived DAGs}} & JSON format & \textcolor{orange!80!black}{CwoM}, \textcolor{orange!80!black}{CwoM} & \textcolor{cyan!80!black}{IND}, \textcolor{cyan!80!black}{IND} & \textcolor{cyan!80!black}{IND}, \textcolor{cyan!80!black}{IND} \\
\multicolumn{2}{c}{} & XML format & \textcolor{violet}{INS}, \textcolor{orange!80!black}{CwoM} & \textcolor{cyan!80!black}{IND}, \textcolor{cyan!80!black}{IND} & \textcolor{cyan!80!black}{IND}, \textcolor{cyan!80!black}{IND} \\
\multicolumn{2}{c}{} & YAML format & \textcolor{orange!80!black}{CwoM}, \textcolor{orange!80!black}{CwoM} & \textcolor{cyan!80!black}{IND}, \textcolor{cyan!80!black}{IND} & \textcolor{cyan!80!black}{IND}, \textcolor{cyan!80!black}{IND} \\ \hline
\end{tabular}%
}
\caption{Discovery of GPT-4.1’s DAGs based on structured output by format-restricting instruction. The $p$-values are reported in parentheses, with 0.000 indicating a value smaller than 0.0005 due to rounding to three decimal places. Each setting shows two DAGs based on $\alpha$ values of 0.05 (first) and 0.1 (second).}
\label{gpt41}
\end{table*}

\begin{table*}[t]
\resizebox{\textwidth}{!}{%
\begin{tabular}{cllccc}
\hline
Hypotheses & Fixed modules & Examined modules & SOT & OpsEval & XCodeEval \\ \hline
\multirow{4}{*}{A} & \multirow{4}{*}{\begin{tabular}[c]{@{}l@{}}Standard \& alternative\\ instruction\end{tabular}} & \begin{tabular}[c]{@{}l@{}}Controlled\\ (w/ standard format)\end{tabular} & 0.988 & 0.305 & 0.854 \\
 &  & \begin{tabular}[c]{@{}l@{}}Intervened\\ (w/ JSON format)\end{tabular} & \begin{tabular}[c]{@{}c@{}}1.000\\ (0.839)\end{tabular} & \begin{tabular}[c]{@{}c@{}}0.310\\ (1.000)\end{tabular} & \begin{tabular}[c]{@{}c@{}}0.858\\ (0.865)\end{tabular} \\
 &  & \begin{tabular}[c]{@{}l@{}}Intervened\\ (w/ XML format)\end{tabular} & \begin{tabular}[c]{@{}c@{}}1.000\\ (0.839)\end{tabular} & \begin{tabular}[c]{@{}c@{}}0.295\\ (1.000)\end{tabular} & \begin{tabular}[c]{@{}c@{}}0.847\\ (0.727)\end{tabular} \\
 &  & \begin{tabular}[c]{@{}l@{}}Intervened\\ (w/ YAML format)\end{tabular} & \begin{tabular}[c]{@{}c@{}}1.000\\ (0.839)\end{tabular} & \begin{tabular}[c]{@{}c@{}}0.315\\ (0.883)\end{tabular} & \begin{tabular}[c]{@{}c@{}}0.862\\ (0.809)\end{tabular} \\ \hline
\multirow{6}{*}{B} & \multirow{2}{*}{Standard \& JSON format} & \begin{tabular}[c]{@{}l@{}}Controlled\\ (w/ standard instruction)\end{tabular} & 1.000 & 0.325 & 0.860 \\
 &  & \begin{tabular}[c]{@{}l@{}}Intervened\\ (w/ alternative instruction)\end{tabular} & \begin{tabular}[c]{@{}c@{}}0.999\\ (0.295)\end{tabular} & \begin{tabular}[c]{@{}c@{}}0.303\\ (0.776)\end{tabular} & \begin{tabular}[c]{@{}c@{}}0.855\\ (0.945)\end{tabular} \\ \cline{2-6} 
 & \multirow{2}{*}{Standard \& XML format} & \begin{tabular}[c]{@{}l@{}}Controlled\\ (w/ standard instruction)\end{tabular} & 1.000 & 0.312 & 0.860 \\
 &  & \begin{tabular}[c]{@{}l@{}}Intervened\\ (w/ alternative instruction)\end{tabular} & \begin{tabular}[c]{@{}c@{}}0.999\\ (0.295)\end{tabular} & \begin{tabular}[c]{@{}c@{}}0.297\\ (0.832)\end{tabular} & \begin{tabular}[c]{@{}c@{}}0.848\\ (0.747)\end{tabular} \\ \cline{2-6} 
 & \multirow{2}{*}{Standard \& YAML format} & \begin{tabular}[c]{@{}l@{}}Controlled\\ (w/ standard instruction)\end{tabular} & 1.000 & 0.325 & 0.865 \\
 &  & \begin{tabular}[c]{@{}l@{}}Intervened\\ (w/ alternative instruction)\end{tabular} & \begin{tabular}[c]{@{}c@{}}0.999\\ (0.295)\end{tabular} & \begin{tabular}[c]{@{}c@{}}0.306\\ (0.809)\end{tabular} & \begin{tabular}[c]{@{}c@{}}0.856\\ (0.928)\end{tabular} \\ \hline
\multirow{3}{*}{C} & \multirow{3}{*}{LLMs' generation} & JSON format & - & - & - \\
 &  & XML format & - & - & - \\
 &  & YAML format & - & - & - \\ \hline
\multicolumn{2}{c}{\multirow{3}{*}{Derived DAGs}} & JSON format & \textcolor{cyan!80!black}{IND}, \textcolor{cyan!80!black}{IND} & \textcolor{cyan!80!black}{IND}, \textcolor{cyan!80!black}{IND} & \textcolor{cyan!80!black}{IND}, \textcolor{cyan!80!black}{IND} \\
\multicolumn{2}{c}{} & XML format & \textcolor{cyan!80!black}{IND}, \textcolor{cyan!80!black}{IND} & \textcolor{cyan!80!black}{IND}, \textcolor{cyan!80!black}{IND} & \textcolor{cyan!80!black}{IND}, \textcolor{cyan!80!black}{IND} \\
\multicolumn{2}{c}{} & YAML format & \textcolor{cyan!80!black}{IND}, \textcolor{cyan!80!black}{IND} & \textcolor{cyan!80!black}{IND}, \textcolor{cyan!80!black}{IND} & \textcolor{cyan!80!black}{IND}, \textcolor{cyan!80!black}{IND} \\ \hline
\end{tabular}%
}
\caption{Discovery of OpenAI-o3’s DAGs based on structured output by format-restricting instruction. The $p$-values are reported in parentheses, with 0.000 indicating a value smaller than 0.0005 due to rounding to three decimal places. Each setting shows two DAGs based on $\alpha$ values of 0.05 (first) and 0.1 (second).}
\label{openaio3}
\end{table*}

\section{Function calling}
\label{function_calling}
The format-restricting instructions are generalizable to all language models. However, this approach embeds both the instruction and output format within the prompt, raising concerns about whether the marginal independence is fully satisfied. To address this, we conducted additional experiments using GPT-4o's function calling to generate JSON outputs based on the function definition in Figure 
~\ref{function_call_definition}. In this setup, the instruction prompt is kept consistent across both structured and unstructured formats, while the structured output format is specified outside the prompt, fully satisfying the marginal independence.

Table \ref{function_call_table} presents the DAGs of GPT-4o under structured output generated via function calling. Consistent with the results in the main text, independence remains the predominant structure among the 16 DAGs. Notably, structured output produced through function calling outperforms that generated via format-restricting instruction in 5 out of 8 datasets. We hypothesize that this improvement stems from further optimization of GPT-4o for generation quality in the function calling setting, and we therefore recommend that industrial practitioners test both approaches before system deployment.

\begin{tcolorbox}[breakable, enhanced]
structured\underline{ }output = [\\%
\{\\%
    \hspace*{1em}"type": "function",\\%
    \hspace*{1em}"function": \{\\%
        \hspace*{2em}"name": "structured\underline{ }output",\\%
        \hspace*{2em}"description": "Output the reasoning steps leading to the final conclusion. Output the final answer, taking into account the reasoning steps. Make sure the answer only contain the final character string.",\\%
        \hspace*{2em}"parameters": \{\\%
            \hspace*{3em}"type": "object",\\%
            \hspace*{3em}"properties": \{\\%
                \hspace*{4em}"reasoning": \{\\%
                    \hspace*{5em}"type": "string",\\%
                    \hspace*{5em}"description": "The reasoning steps leading to the final conclusion.",\\%
                \},\\%
                \hspace*{4em}"answer": \{\\%
                    \hspace*{5em}"type": "string",\\%
                    \hspace*{5em}"description": "The final answer, taking into account the reasoning steps. Make sure the answer only contain the final character string.",\\%
                \hspace*{4em}\},\\%
            \hspace*{3em}\},\\%
            \hspace*{3em}"required": ["reasoning", "answer"]\\%
        \hspace*{2em}\}\\%
    \hspace*{1em}\}\\%
\}\\%
]
\end{tcolorbox}
\noindent\begin{minipage}{\columnwidth}
\captionof{figure}{Function definition for GPT-4o JSON output.}\label{function_call_definition}
\end{minipage}

\section{General-purpose and reasoning models}
\label{appendix_41_o3}
Most experiments in the main text use the general-purpose GPT-4o as the backbone. To further examine the impact of structured output on the latest general-purpose and reasoning models, we additionally conduct experiments on GPT-4.1 and OpenAI-o3. Tables \ref{gpt41} and \ref{openaio3} presents the DAGs of these two models released after GPT-4o. GPT-4.1, the general-purpose successor, exhibits patterns similar to GPT-4o, indicating that its performance on the SOT dataset is statistically significantly affected by structured output formats. In contrast, OpenAI-o3, a high-performance reasoning model, achieves perfect accuracy on the SOT dataset across all settings except for a single case involving intervened instruction under the unstructured format. Consequently, its DAGs and p-values remain stable across structured formats. On the other two datasets of OpsEval and XCodeEval, the content quality of OpenAI-o3 is also not statistically affected by output format, highlighting a complementary advantage of reasoning-intensive models over general-purpose models.

\section{Small language models}
\label{open_source}
In the industry context, we are granted access to a limited number of models that do not include all prevalent LLMs and their latest versions. Also, the computational resource is restricted to one NVIDIA A100 40GB device and therefore we need to use relatively small models for inference. We use the format-restricting instruction strategy to prompt LLMs to generate structured output.

We evaluate GPT-oss-20B, Llama-3.1-8B, Gemma-2-9B, and Phi-3.5-mini. While we acknowledge that these may not be state-of-the-art (SOTA) models due to previously mentioned limitations, they remain frequently downloaded according to Ollama as of August 2026, and therefore we believe they are representative. Due to computational resource constraints, the maximum token generation is limited to 1024. While this is sufficient for the three non-reasoning models, it may restrict GPT-oss-20B and affect its performance.

\begin{figure*}[h]
    \centering
\includegraphics[width=\textwidth]{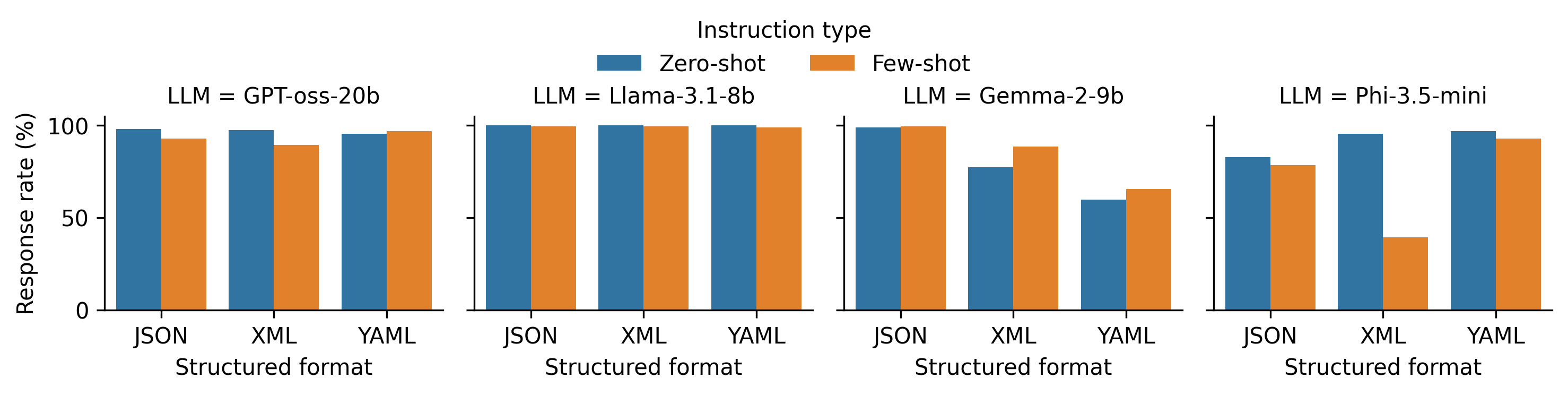}
    \caption{Successful response rate of different SLMs under zero-shot and few-shot Prompts.}
    \label{response_rate}
\end{figure*}

Considering the limited capabilities of SLMs compared to GPT-4o, their ability to follow instructions for structured output formats is correspondingly weaker, and the zero-shot prompts used for GPT-4o may not be sufficient. To address this, we apply few-shot prompts inspired by LangChain through providing both positive and negative examples of the specified format to guide the models' responses (Figures \ref{few_shot_json_format}, \ref{few_shot_xml_format}, and \ref{few_shot_yaml_format}). Figure \ref{response_rate} shows that few-shot prompting improves response rates for YAML with GPT-oss-20B, and for both XML and YAML with Gemma-2-9B; therefore, we apply few-shot prompting in these scenarios while retaining zero-shot prompting for the others.

\begin{tcolorbox}[breakable, enhanced]
Few-shot JSON format prompt\\
The output should be formatted as a raw JSON instance that conforms to the JSON schema below without code fences.\\
As an example, for the schema\\
\{\\
"reasoning": "reasoning string",\\
"answer": "answer string"\\
\}\\
the object\\
\{\\
"reasoning": "bar",\\
"answer": "baz"\\
\}\\
is a well-formatted instance of the schema.\\
The object\\
\{\\
"reasoning": "bar"\\
"answer": "baz"\\
]\}\\
is not well-formatted.\\
Here is the output schema:\\
\{\\
"reasoning": "The reasoning steps leading to the final conclusion.",\\
"answer": "The final answer, taking into account the reasoning steps."\\
\}
\end{tcolorbox}
\noindent\begin{minipage}{\columnwidth}
\captionof{figure}{Few-shot JSON prompt for SLMs.}\label{few_shot_json_format}
\end{minipage}

\begin{tcolorbox}[breakable, enhanced]
Few-shot XML format prompt\\
The output should be formatted as a raw XML instance that conforms to the XML schema below without code fences.\\
As an example, for the schema\\
<root>\\
<reasoning>[reasoning string]</reasoning>\\
<answer>[answer string]<answer>\\
</root>\\
the object\\
<root>\\
<reasoning>bar</reasoning>\\
<answer>baz<answer>\\
</root>\\
is a well-formatted instance of the schema.\\
The object\\
<root>\\
<reasoning>bar</reasoning>\\
<answer>baz<answer>\\
is not well-formatted.\\
Here is the output schema:\\
<root>\\
<reasoning>[The reasoning steps leading to the final conclusion.]</reasoning>\\
<answer>[The final answer, taking into account the reasoning steps.]<answer>\\
</root>
\end{tcolorbox}
\noindent\begin{minipage}{\columnwidth}
\captionof{figure}{Few-shot XML prompt for SLMs.}\label{few_shot_xml_format}
\end{minipage}

\begin{tcolorbox}[breakable, enhanced]
Few-shot YAML format prompt\\
The output should be formatted as a raw YAML instance that conforms to the YAML schema below without code fences.\\
As an example, for the schema\\
reasoning: |\\
\hspace*{1em}"reasoning string"\\
answer: "answer string"\\
the object\\
reasoning: "bar"\\
answer: "baz"\\
is a well-formatted instance of the schema.\\
The object\\
reasoning: "bar"]\}\\
is not well-formatted.\\
Here is the output schema:\\
reasoning: |\\
\hspace*{1em}"<The reasoning steps leading to the final conclusion.>"\\
answer: "<The final answer, taking into account the reasoning steps.>"
\end{tcolorbox}
\noindent\begin{minipage}{\columnwidth}
\captionof{figure}{Few-shot YAML prompt for SLMs.}\label{few_shot_yaml_format}
\end{minipage}

\begin{figure*}[t]
    \centering
\includegraphics[width=\textwidth]{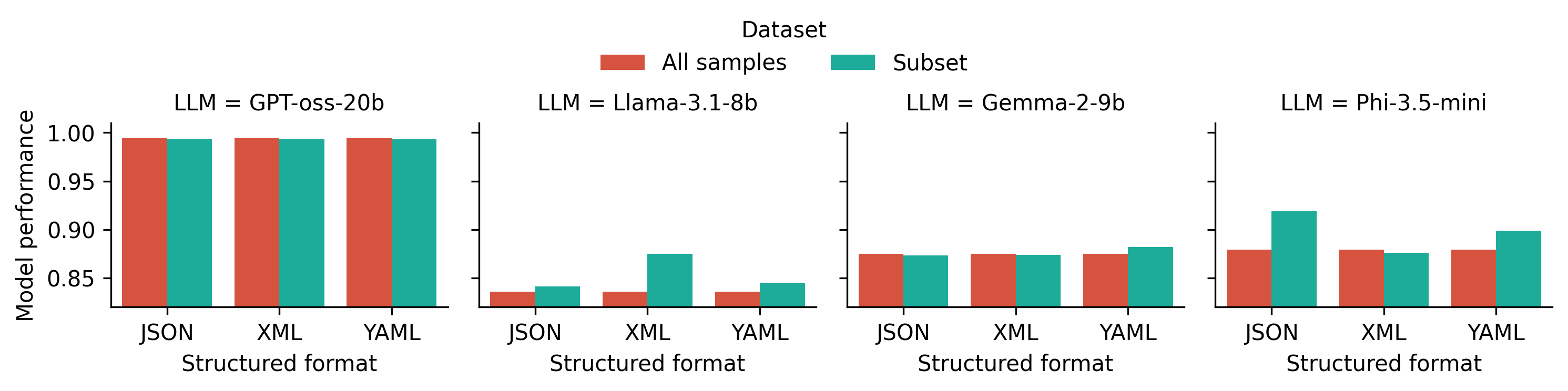}
    \caption{Comparison of unstructured generation on entire dataset and subsets of successful responses.}
    \label{subset_performance}
\end{figure*}

\begin{table*}[t]
\resizebox{\textwidth}{!}{%
\begin{tabular}{cllcccc}
\hline
Hypotheses & Fixed modules & Examined modules & GPT-oss-20B & Llama-3.1-8B & Gemma-2-9B & Phi-3.5-mini \\ \hline
\multirow{6}{*}{A} & \multirow{6}{*}{\begin{tabular}[c]{@{}l@{}}Standard \& alternative\\ instruction\end{tabular}} & \begin{tabular}[c]{@{}l@{}}Controlled\\ (w/ standard format)\end{tabular} & 0.993 & 0.841 & 0.873 & 0.919 \\ & & \begin{tabular}[c]{@{}l@{}}Intervened\\ (w/ JSON format)\end{tabular}             & \begin{tabular}[c]{@{}c@{}}0.989\\ (0.839)\end{tabular} & \begin{tabular}[c]{@{}c@{}}0.603\\ (0.000)\end{tabular} & \begin{tabular}[c]{@{}c@{}}0.849\\ (0.241)\end{tabular} & \begin{tabular}[c]{@{}c@{}}0.889\\ (0.130)\end{tabular} \\ \cline{3-7} & & \begin{tabular}[c]{@{}l@{}}Controlled\\ (w/ standard format)\end{tabular}         & 0.993 & 0.875 & 0.874 & 0.876 \\ & & \begin{tabular}[c]{@{}l@{}}Intervened\\ (w/ XML format)\end{tabular}              & \begin{tabular}[c]{@{}c@{}}0.989\\ (0.839)\end{tabular} & \begin{tabular}[c]{@{}c@{}}0.882\\ (0.042)\end{tabular} & \begin{tabular}[c]{@{}c@{}}0.877\\ (0.285)\end{tabular} & \begin{tabular}[c]{@{}c@{}}0.860\\ (0.296)\end{tabular} \\ \cline{3-7} & & \begin{tabular}[c]{@{}l@{}}Controlled\\ (w/ standard format)\end{tabular}         & 0.993 & 0.845 & 0.882 & 0.899 \\ & & \begin{tabular}[c]{@{}l@{}}Intervened\\ (w/ YAML format)\end{tabular}             & \begin{tabular}[c]{@{}c@{}}0.985\\ (0.418)\end{tabular} & \begin{tabular}[c]{@{}c@{}}0.799\\ (0.003)\end{tabular} & \begin{tabular}[c]{@{}c@{}}0.880\\ (0.594)\end{tabular} & \begin{tabular}[c]{@{}c@{}}0.861\\ (0.066)\end{tabular} \\ \hline
\multirow{6}{*}{B} & \multirow{2}{*}{Standard \& JSON format} & \begin{tabular}[c]{@{}l@{}}Controlled\\ (w/ standard instruction)\end{tabular}    & 0.986 & 0.757 & 0.875 & 0.906 \\ & & \begin{tabular}[c]{@{}l@{}}Intervened\\ (w/ alternative instruction)\end{tabular} & \begin{tabular}[c]{@{}c@{}}0.992\\ (0.162)\end{tabular} & \begin{tabular}[c]{@{}c@{}}0.714\\ (0.000)\end{tabular} & \begin{tabular}[c]{@{}c@{}}0.858\\ (0.001)\end{tabular} & \begin{tabular}[c]{@{}c@{}}0.903\\ (0.231)\end{tabular} \\ \cline{2-7} & \multirow{2}{*}{Standard \& XML format} & \begin{tabular}[c]{@{}l@{}}Controlled\\ (w/ standard instruction)\end{tabular}    & 0.983 & 0.899 & 0.887 & 0.874 \\ & & \begin{tabular}[c]{@{}l@{}}Intervened\\ (w/ alternative instruction)\end{tabular} & \begin{tabular}[c]{@{}c@{}}0.993\\ (0.162)\end{tabular} & \begin{tabular}[c]{@{}c@{}}0.873\\ (0.000)\end{tabular} & \begin{tabular}[c]{@{}c@{}}0.873\\ (0.053)\end{tabular} & \begin{tabular}[c]{@{}c@{}}0.866\\ (0.174)\end{tabular} \\ \cline{2-7} & \multirow{2}{*}{Standard \& YAML format} & \begin{tabular}[c]{@{}l@{}}Controlled\\ (w/ standard instruction)\end{tabular}    & 0.986  & 0.831 & 0.889 & 0.891 \\ &                                                                                                & \begin{tabular}[c]{@{}l@{}}Intervened\\ (w/ alternative instruction)\end{tabular} & \begin{tabular}[c]{@{}c@{}}0.990\\ (0.063)\end{tabular} & \begin{tabular}[c]{@{}c@{}}0.820\\ (0.000)\end{tabular} & \begin{tabular}[c]{@{}c@{}}0.879\\ (0.000)\end{tabular} & \begin{tabular}[c]{@{}c@{}}0.877\\ (0.189)\end{tabular} \\ \hline
\multirow{3}{*}{C} & \multirow{3}{*}{LLMs' generation}  & JSON format & - & 0.059 & -    & -  \\  & & XML format & -  & 0.973  & -  & - \\ & & YAML format     & -  & 0.790  & -  &  \\ \hline
\multicolumn{2}{c}{\multirow{3}{*}{Derived DAGs}}  & JSON format & \textcolor{cyan!80!black}{IND}, \textcolor{cyan!80!black}{IND} & \textcolor{orange!80!black}{CwoM}, \textcolor {red!90!black}{CwM}  & \textcolor{violet}{INS}, \textcolor{violet}{INS}  & \textcolor{cyan!80!black}{IND}, \textcolor{cyan!80!black}{IND}   \\
\multicolumn{2}{c}{}                      & XML format   & \textcolor{cyan!80!black}{IND}, \textcolor{cyan!80!black}{IND}  & \textcolor{orange!80!black}{CwoM}, \textcolor{orange!80!black}{CwoM}  & \textcolor{cyan!80!black}{IND}, \textcolor{violet}{INS} & \textcolor{cyan!80!black}{IND}, \textcolor{cyan!80!black}{IND} \\
\multicolumn{2}{c}{}                       & YAML format   & \textcolor{cyan!80!black}{IND}, \textcolor{violet}{INS}  & \textcolor{orange!80!black}{CwoM}, \textcolor{orange!80!black}{CwoM}  & \textcolor{violet}{INS}, \textcolor{violet}{INS}  & \textcolor{cyan!80!black}{IND}, \textcolor{green!60!black}{FMT}  \\ \hline
\end{tabular}%
}
\caption{Discovery of SLMs' DAGs based on structured output by format-restricting instruction. Only cases with successful structured format generation are included. The $p$-values are reported in parentheses, with 0.000 indicating a value smaller than 0.0005 due to rounding to three decimal places.}
\label{oa_causal_discovery}
\end{table*}

\begin{table*}[b]
\resizebox{\textwidth}{!}{%
\begin{tabular}{cllcccc}
\hline
Hypotheses& Fixed modules& Examined modules & GPT-oss-20B & Llama-3.1-8B & Gemma-2-9B    & Phi-3.5-mini \\ \hline
\multirow{6}{*}{A} & \multirow{6}{*}{\begin{tabular}[c]{@{}l@{}}Standard \& alternative\\ instruction\end{tabular}} & \begin{tabular}[c]{@{}l@{}}Controlled\\ (w/ standard format)\end{tabular}         & 0.962 & 0.836 & 0.875 & 0.879 \\  &          & \begin{tabular}[c]{@{}l@{}}Intervened\\ (w/ JSON format)\end{tabular}             & \begin{tabular}[c]{@{}c@{}}0.954\\ (0.550)\end{tabular} & \begin{tabular}[c]{@{}c@{}}0.586\\ (0.000)\end{tabular} & \begin{tabular}[c]{@{}c@{}}0.847\\ (0.165)\end{tabular} & \begin{tabular}[c]{@{}c@{}}0.711\\ (0.000)\end{tabular} \\ \cline{3-7} &   & \begin{tabular}[c]{@{}l@{}}Controlled\\ (w/ standard format)\end{tabular}         & 0.962         & 0.836  & 0.875  & 0.879 \\ &  & \begin{tabular}[c]{@{}l@{}}Intervened\\ (w/ XML format)\end{tabular}              & \begin{tabular}[c]{@{}c@{}}0.957\\ (0.210)\end{tabular} & \begin{tabular}[c]{@{}c@{}}0.833\\ (0.021)\end{tabular} & \begin{tabular}[c]{@{}c@{}}0.875\\ (0.318)\end{tabular} & \begin{tabular}[c]{@{}c@{}}0.787\\ (0.000)\end{tabular} \\ \cline{3-7}  &  & \begin{tabular}[c]{@{}l@{}}Controlled\\ (w/ standard format)\end{tabular}         & 0.962         & 0.836  & 0.875  & 0.879 \\ &  & \begin{tabular}[c]{@{}l@{}}Intervened\\ (w/ YAML format)\end{tabular}             & \begin{tabular}[c]{@{}c@{}}0.949\\ (0.291)\end{tabular} & \begin{tabular}[c]{@{}c@{}}0.788\\ (0.002)\end{tabular} & \begin{tabular}[c]{@{}c@{}}0.851\\ (0.081)\end{tabular} & \begin{tabular}[c]{@{}c@{}}0.799\\ (0.000)\end{tabular} \\ \hline
\multirow{6}{*}{B} & \multirow{2}{*}{Standard \& JSON format}                    & \begin{tabular}[c]{@{}l@{}}Controlled\\ (w/ standard instruction)\end{tabular}    & 0.958  & 0.750  & 0.877 & 0.802 \\ &         & \begin{tabular}[c]{@{}l@{}}Intervened\\ (w/ alternative instruction)\end{tabular} & \begin{tabular}[c]{@{}c@{}}0.958\\ (0.563)\end{tabular} & \begin{tabular}[c]{@{}c@{}}0.701\\ (0.000)\end{tabular} & \begin{tabular}[c]{@{}c@{}}0.857\\ (0.001)\end{tabular} & \begin{tabular}[c]{@{}c@{}}0.793\\ (0.152)\end{tabular} \\ \cline{2-7}  & \multirow{2}{*}{Standard \& XML format}  & \begin{tabular}[c]{@{}l@{}}Controlled\\ (w/ standard instruction)\end{tabular}    & 0.958         & 0.865 & 0.887  & 0.863 \\ & & \begin{tabular}[c]{@{}l@{}}Intervened\\ (w/ alternative instruction)\end{tabular} & \begin{tabular}[c]{@{}c@{}}0.960\\ (0.070)\end{tabular} & \begin{tabular}[c]{@{}c@{}}0.827\\ (0.000)\end{tabular} & \begin{tabular}[c]{@{}c@{}}0.872\\ (0.037)\end{tabular} & \begin{tabular}[c]{@{}c@{}}0.826\\ (0.000)\end{tabular} \\ \cline{2-7} 
 & \multirow{2}{*}{Standard \& YAML format}  & \begin{tabular}[c]{@{}l@{}}Controlled\\ (w/ standard instruction)\end{tabular}    & 0.958 & 0.823 & 0.855 & 0.855 \\
 &  & \begin{tabular}[c]{@{}l@{}}Intervened\\ (w/ alternative instruction)\end{tabular} & \begin{tabular}[c]{@{}c@{}}0.955\\ (0.375)\end{tabular} & \begin{tabular}[c]{@{}c@{}}0.809\\ (0.000)\end{tabular} & \begin{tabular}[c]{@{}c@{}}0.865\\ (0.000)\end{tabular} & \begin{tabular}[c]{@{}c@{}}0.835\\ (0.012)\end{tabular} \\ \hline
\multirow{3}{*}{C} & \multirow{3}{*}{LLMs' generation}   & JSON format & -  & 0.040     & -  & -  \\ &  & XML format & -            & 0.987 & - & 0.646 \\ & & YAML format      & - & 0.798 & 0.940 & 0.792 \\ \hline
\multicolumn{2}{c}{\multirow{3}{*}{Derived DAGs}} & JSON format & \textcolor{cyan!80!black}{IND}, \textcolor{cyan!80!black}{IND}   & \textcolor {red!90!black}{CwM}, \textcolor {red!90!black}{CwM}  & \textcolor{violet}{INS}, \textcolor{violet}{INS}   & \textcolor{green!60!black}{FMT}, \textcolor{green!60!black}{FMT}  \\
\multicolumn{2}{c}{}                          & XML format  & \textcolor{cyan!80!black}{IND}, \textcolor{violet}{INS}  & \textcolor{orange!80!black}{CwoM}, \textcolor{orange!80!black}{CwoM}   & \textcolor{violet}{INS}, \textcolor{violet}{INS}  & \textcolor{orange!80!black}{CwoM}, \textcolor{orange!80!black}{CwoM}  \\
\multicolumn{2}{c}{}                         & YAML format & \textcolor{cyan!80!black}{IND}, \textcolor{cyan!80!black}{IND} & \textcolor{orange!80!black}{CwoM}, \textcolor{orange!80!black}{CwoM} & \textcolor{violet}{INS}, \textcolor{orange!80!black}{CwoM}  & \textcolor{orange!80!black}{CwoM}, \textcolor{orange!80!black}{CwoM}  \\ \hline
\end{tabular}%
}
\caption{Discovery of SLMs' DAGs based on structured output by format-restricting instruction. All cases are included in the causal inference and cases with incorrect format are considered as false. The $p$-values are reported in parentheses, with 0.000 indicating a value smaller than 0.0005 due to rounding to three decimal places.}
\label{slm_full}
\end{table*}

To enable compatibility of statistical tests and causal discovery, we retain, for each model, only samples with successful responses in structured output formats across all instruction interventions. This way may introduce post-treatment selection bias, since models may more easily produce structured outputs for simpler questions, potentially overestimating accuracy. Alternatively, we could treat unsuccessful responses as incorrect answers, but this may underestimate the models' true performance and exaggerate DAGs.

First, we conduct causal discovery based on the selected samples with successful responses across diverse interventions, Table \ref{oa_causal_discovery} presents the causal discovery process for SLMs. In SLMs other than GPT-oss-20B, the standard instruction format consistently outperforms alternative instructions, suggesting that a persona well-aligned with the target task is crucial. Second, in ten out of twelve scenarios, structured output formats degrade model performance, highlighting the limited ability of SLMs to maintain output quality when constrained by structured formats. This suggests that, for practical applications, SLMs generating structured outputs should be benchmarked against SLMs producing unstructured outputs paired with external parsers. Third, under a 0.1 significance level, we detect the causal structure of CwM in JSON generation of Llama-3.1-8B, a pattern not found in GPT-4o, indicating that spurious correlations between instruction style and output format may emerge under certain conditions. Lastly, since the causal analysis is conducted on subsets with successful structured responses, we compare model performance of unstructured generation on these subsets versus the full dataset. Given the near-perfect performance of GPT-oss-20B, we exclude its three scenarios from this analytical perspective. As shown in Figure \ref{subset_performance}, in six of nine scenarios, SLMs perform better on the subset with valid structured outputs. We hypothesize two possible reasons: (1) models may more easily produce structured outputs on simpler questions, and (2) generating a structured format may correlate with overall higher generation quality. These hypotheses warrant future investigation.

Alternatively, we treat cases in which the target structured format is not correctly generated as false cases and perform causal discovery on the full set. Table~\ref{slm_full} presents the results. Seven DAGs previously classified as independence are now identified as exhibiting statistically significant impacts. Consistent with our expectations, including all cases in the causal inference helps mitigate post-selection bias, whereas treating unsuccessful structured format generations as false cases may underestimate model performance due to the inherent difficulties of format generation, potentially resulting in more statistically significant relationships and DAGs with more complex structures.

\section{Additional interventions}
\label{additional_interventions}
For the interventions discussed in the main text, the output format intervention comprehensively covers three structured formats, while the instruction intervention follows \citet{bao-etal-2025-likely} by employing predefined paraphrased system prompts, with the context description and problem formulation held constant. Here, we present additional intervention experiments targeting the context description and problem formulation modules, conducted on the first 50 samples from GSM8K. We first prompt GPT-4o to generate three variants of the original context description and problem formulation (Figure \ref{instruction_intervention_prompt}). These variants are then manually reviewed and refined to ensure they preserve the original reasoning steps and final answers. Figure \ref{gsm8k_intervention} provides an illustrative example from GSM8K, highlighting the variations introduced through interventions on the context description and problem formulation.


Table \ref{table_add_intervention} presents the results of GPT-4o's DAG discovery under output format intervention, which is same as the main text, and instruction intervention, which modifies the context description and problem formulation. The identified DAGs remain consistent with those in the main text. However, we observe universal performance drops after intervention, even when the reasoning steps and final answers remain the same and only the expression is modified. This suggests that current LLMs may be overfitting to existing public benchmarks, underscoring the need for new benchmarks (e.g. ELLC) that are unseen to the models \citep{pmlr-v235-chen24i,mirzadeh2025gsmsymbolic}.

\begin{tcolorbox}[breakable, enhanced]
You are given a math problem that includes a context and a question. Your task is to rephrase the entire problem to make it sound different in wording and style, while keeping the following strictly unchanged:
meaning and intent of the problem;
mathematical structure and calculation steps required to solve it;
final numerical or symbolic answer to the problem.
Avoid introducing or removing any information. You may change names, objects, settings, or reword sentences, but do not alter quantities, relationships, or operations.\\
Original problem:
\end{tcolorbox}
\noindent\begin{minipage}{\columnwidth}
\captionof{figure}{Intervention prompt for GSM8K variants.}\label{instruction_intervention_prompt}
\end{minipage}

\begin{tcolorbox}[breakable, enhanced]
Original instruction\\
John plans to sell all his toys and use the money to buy video games. He has 13 lego sets and he sells them for \$15 each. He ends up buying 8 video games for \$20 each and has \$5 left. How many lego sets does he still have?\\
\medskip
\noindent\dotfill\\
\medskip
Intervened instruction 0\\
Mark intends to sell his entire collection of model cars to purchase books. He owns 13 model cars, each sold for \$15. Eventually, he buys 8 books at \$20 each and retains \$5. How many model cars does he still possess?\\
\medskip
\noindent\dotfill\\
\medskip
Intervened instruction 1\\
Mike wants to exchange his toy collection for some video games. He possesses 13 toy car sets, and each one is sold for \$15. Mike buys 8 video games priced at \$20 each, and he has \$5 remaining after his purchase. How many toy car sets does Mike have left?\\
\medskip
\noindent\dotfill\\
Intervened instruction 2\\
John is looking to sell his collection of toys to purchase some video games. He owns a total of 13 lego sets, each of which he sells for \$15. He buys 8 video games at the price of \$20 each and after the transaction has \$5 remaining. How many lego sets does he have left?
\end{tcolorbox}
\noindent\begin{minipage}{\columnwidth}
\captionof{figure}{Interventions of context description and problem formulation on the first sample from GSM8K.}\label{gsm8k_intervention}
\end{minipage}

\begin{table*}[t]
\centering
\small
\begin{tabular}{cllc}
\hline
Hypotheses         & Fixed modules & Examined modules     & GSM8K \\ \hline
\multirow{4}{*}{A} & \multirow{4}{*}{\begin{tabular}[c]{@{}l@{}}Standard \& alternative\\ instruction\end{tabular}} & \begin{tabular}[c]{@{}l@{}}Controlled\\ (w/ standard format)\end{tabular}         & 0.930      \\&       & \begin{tabular}[c]{@{}l@{}}Intervened\\ (w/ JSON format)\end{tabular}             & \begin{tabular}[c]{@{}c@{}}0.950\\ (0.404)\end{tabular} \\
&                           & \begin{tabular}[c]{@{}l@{}}Intervened\\ (w/ XML format)\end{tabular}              & \begin{tabular}[c]{@{}c@{}}0.950\\ (0.625)\end{tabular} \\
&                         & \begin{tabular}[c]{@{}l@{}}Intervened\\ (w/ YAML format)\end{tabular}             & \begin{tabular}[c]{@{}c@{}}0.940\\ (0.859)\end{tabular} \\ \hline
\multirow{6}{*}{B} & \multirow{2}{*}{Standard \& JSON format}                   & \begin{tabular}[c]{@{}l@{}}Controlled\\ (w/ standard instruction)\end{tabular}    & 0.950                  \\ &  & \begin{tabular}[c]{@{}l@{}}Intervened\\ (w/ alternative instruction)\end{tabular} & \begin{tabular}[c]{@{}c@{}}0.937\\ (0.288)\end{tabular} \\ \cline{2-4} 
& \multirow{2}{*}{Standard \& XML format}                    & \begin{tabular}[c]{@{}l@{}}Controlled\\ (w/ standard instruction)\end{tabular}    & 0.960       \\ &              & \begin{tabular}[c]{@{}l@{}}Intervened\\ (w/ alternative instruction)\end{tabular} & \begin{tabular}[c]{@{}c@{}}0.933\\ (0.468)\end{tabular} \\ \cline{2-4} 
& \multirow{2}{*}{Standard \& YAML format}                   & \begin{tabular}[c]{@{}l@{}}Controlled\\ (w/ standard instruction)\end{tabular}    & 0.960            \\&           & \begin{tabular}[c]{@{}l@{}}Intervened\\ (w/ alternative instruction)\end{tabular} & \begin{tabular}[c]{@{}c@{}}0.927\\ (0.315)\end{tabular} \\ \hline
\multirow{3}{*}{C} & \multirow{3}{*}{LLMs' generation}                    & JSON format      & -       \\
&                              & XML format                   & -                         \\
&                              & YAML format                  & -                   \\ \hline
\multicolumn{2}{c}{\multirow{3}{*}{Derived DAGs}}             & JSON format                  & \textcolor{cyan!80!black}{IND}, \textcolor{cyan!80!black}{IND}                   \\
\multicolumn{2}{c}{}           & XML format                   & \textcolor{cyan!80!black}{IND}, \textcolor{cyan!80!black}{IND}                  \\
\multicolumn{2}{c}{}           & YAML format                  & \textcolor{cyan!80!black}{IND}, \textcolor{cyan!80!black}{IND}            \\ \hline
\end{tabular}%
\caption{Discovery of GPT-4o's DAGs under instruction intervention in context description and problem formulation. Structured output is achieved by format-restricting instruction.}
\label{table_add_intervention}
\end{table*}

\begin{table*}[]
\centering
\small
\begin{tabular}{cllccc}
\hline
Hypotheses         & Fixed modules                          & Examined modules               & API calling 0                  & API calling 1                  & API calling 2                  \\ \hline
\multirow{4}{*}{A} & \multirow{4}{*}{\begin{tabular}[c]{@{}l@{}}Standard \& alternative\\ instruction\end{tabular}} & \begin{tabular}[c]{@{}l@{}}Controlled\\ (w/ standard format)\end{tabular}    & 0.693                          & 0.694                          & 0.685                          \\
&                                & \begin{tabular}[c]{@{}l@{}}Intervened\\ (w/ JSON format)\end{tabular}             & \begin{tabular}[c]{@{}c@{}}0.700\\ (0.582)\end{tabular} & \begin{tabular}[c]{@{}c@{}}0.699\\ (0.700)\end{tabular} & \begin{tabular}[c]{@{}c@{}}0.697\\ (0.426)\end{tabular} \\
&                                & \begin{tabular}[c]{@{}l@{}}Intervened\\ (w/ XML format)\end{tabular}              & \begin{tabular}[c]{@{}c@{}}0.689\\ (0.224)\end{tabular} & \begin{tabular}[c]{@{}c@{}}0.689\\ (0.419)\end{tabular} & \begin{tabular}[c]{@{}c@{}}0.688\\ (0.242)\end{tabular} \\
&                                & \begin{tabular}[c]{@{}l@{}}Intervened\\ (w/ YAML format)\end{tabular}             & \begin{tabular}[c]{@{}c@{}}0.713\\ (0.287)\end{tabular} & \begin{tabular}[c]{@{}c@{}}0.719\\ (0.385)\end{tabular} & \begin{tabular}[c]{@{}c@{}}0.706\\ (0.274)\end{tabular} \\ \hline
\multirow{6}{*}{B} & \multirow{2}{*}{Standard \& JSON format}      & \begin{tabular}[c]{@{}l@{}}Controlled\\ (w/ standard instruction)\end{tabular}    & 0.698                            & 0.701                          & 0.705                          \\
&                                & \begin{tabular}[c]{@{}l@{}}Intervened\\ (w/ alternative instruction)\end{tabular} & \begin{tabular}[c]{@{}c@{}}0.696\\ (0.556)\end{tabular} & \begin{tabular}[c]{@{}c@{}}0.695\\ (0.770)\end{tabular} & \begin{tabular}[c]{@{}c@{}}0.688\\ (0.358)\end{tabular} \\ \cline{2-6} 
& \multirow{2}{*}{Standard \& XML format}                                               & \begin{tabular}[c]{@{}l@{}}Controlled\\ (w/ standard instruction)\end{tabular}    & 0.772                            & 0.726                          & 0.722                          \\
&                                & \begin{tabular}[c]{@{}l@{}}Intervened\\ (w/ alternative instruction)\end{tabular} & \begin{tabular}[c]{@{}c@{}}0.683\\ (0.175)\end{tabular} & \begin{tabular}[c]{@{}c@{}}0.683\\ (0.394)\end{tabular} & \begin{tabular}[c]{@{}c@{}}0.677\\ (0.107)\end{tabular} \\ \cline{2-6} 
& \multirow{2}{*}{Standard \& YAML format}                                                       & \begin{tabular}[c]{@{}l@{}}Controlled\\ (w/ standard instruction)\end{tabular}    & 0.701                            & 0.712                          & 0.712                          \\
&                                & \begin{tabular}[c]{@{}l@{}}Intervened\\ (w/ alternative instruction)\end{tabular} & \begin{tabular}[c]{@{}c@{}}0.703\\ (0.287)\end{tabular} & \begin{tabular}[c]{@{}c@{}}0.706\\ (0.452)\end{tabular} & \begin{tabular}[c]{@{}c@{}}0.691\\ (0.109)\end{tabular} \\ \hline
\multirow{3}{*}{C} & \multirow{3}{*}{LLMs' generation}            & JSON format   & -                            & -   & -           \\
&                                & XML format                       & -              & -                      & -                              \\
&                                & YAML format                      & -              & -                      & -                              \\ \hline
\multicolumn{2}{c}{\multirow{3}{*}{Derived DAGs}}               & JSON format & \textcolor{cyan!80!black}{IND}, \textcolor{cyan!80!black}{IND}                   & \textcolor{cyan!80!black}{IND}, \textcolor{cyan!80!black}{IND}                         & \textcolor{cyan!80!black}{IND}, \textcolor{cyan!80!black}{IND}                       \\
\multicolumn{2}{c}{}             & XML format    & \textcolor{cyan!80!black}{IND}, \textcolor{cyan!80!black}{IND}                     & \textcolor{cyan!80!black}{IND}, \textcolor{cyan!80!black}{IND}                         & \textcolor{cyan!80!black}{IND}, \textcolor{cyan!80!black}{IND}                               \\
\multicolumn{2}{c}{}             & YAML format                    & \textcolor{cyan!80!black}{IND}, \textcolor{cyan!80!black}{IND}                       & \textcolor{cyan!80!black}{IND}, \textcolor{cyan!80!black}{IND}                       & \textcolor{cyan!80!black}{IND}, \textcolor{cyan!80!black}{IND}                       \\ \hline
\end{tabular}%
\caption{Discovery of GPT-4o's DAGs across multiple API calls on GCF data. Structured output is achieved by format-restricting instruction. Each setting shows two DAGs based on $\alpha$ values of 0.05 (first) and 0.1 (second).}
\label{table_multi_trial}
\end{table*}

\section{Multiple trials}
\label{multiple_trials}

\begin{figure}[]
    \centering
\includegraphics[width=\columnwidth]{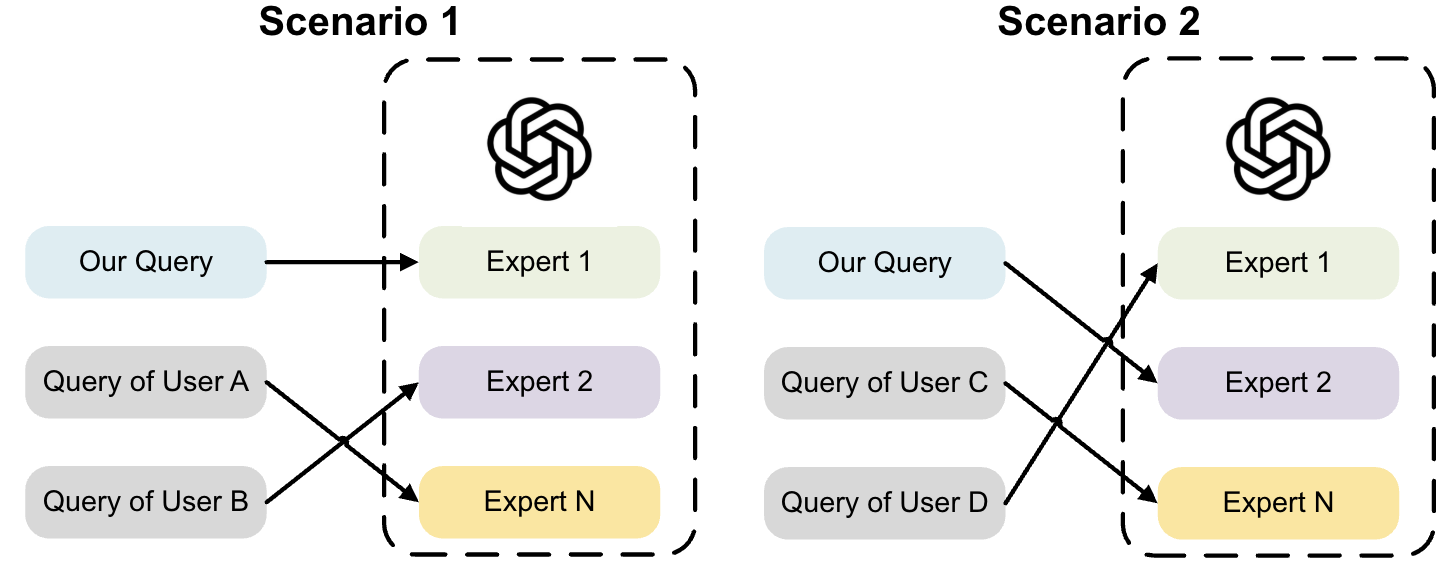}
    \caption{Schematic plot of the impact of MoE architecture on the generation of proprietary LLMs' services.}
    \label{gpt_moe}
\end{figure}

We acknowledge that even under a temperature setting of 0, proprietary GPT-4o, as a highly complex system \citep{cuda}, is subject to numerous sources of variability that influence its outputs. These include computational non-determinism, ties in token probability distributions, diverse backend versions, and dynamic architectures such as mixture-of-experts (Figure \ref{gpt_moe}). As a result, both the numerical outputs and the inferred causal structures may vary across API calls. To mitigate this limitation, this section aggregates results from multiple API calls. Table \ref{table_multi_trial} present results for GCF, as its relatively short token lengths make API costs more manageable. The variation highlights the limitations of conclusive statements regarding structured outputs' impact in prior studies and reinforces our opinion: the influence of structured output formats is nuanced and heterogeneous. It varies across different tasks and schemas, and even across API calls for the same task under an identical schema.

From the perspective of hypothesis testing and causal inference, the results from the second and third API calls are consistent with those of the first trial. However, it is worth noting that in the third API call, the $p$-values for the instruction intervention in the XML and YAML formats are slightly above $\alpha$ of 0.1. While still non-significant, these results suggest the potential for variation in hypothesis test outcomes and inferred causal structures as the number of trials increases, reflecting the inherent variance in LLMs \citep{brown2024large}. Given the increasing number of statistical tests conducted across multiple trials, the inflation of Type I error must be considered, and the Bonferroni correction should be applied.




\section{ELLC benchmark on complex tasks}
\label{ellc_variants}
We extended the LLC dataset to develop the Extended Last Letter Concatenation (or ELLC) dataset, which introduces additional complexity in both symbolic and linguistic reasoning via an evolutionary strategy \citep{li-etal-2024-towards-verifiable}. Symbolic reasoning is enhanced by increasing the number of words and requiring extraction from more challenging positions, such as middle letters. Linguistic reasoning adds a further step: forming common English words from the concatenated letters, thus assessing linguistic capabilities.

The basic task consists of four words, requiring the model to extract the last letter from each word and determine the correct sequence of the extracted letters to form a valid English word. This task can be extended along both reasoning dimensions: symbolic transformation and linguistic reasoning.

For symbolic transformation, task complexity is increased by introducing six words and specifying a more challenging extraction position of the middle letter of each word (Middle Letter Concatenation, or MLC). For linguistic reasoning, the basic question requires composing a single correct answer from all options, while the more advanced question instructs LLMs to enumerate all correct answers. Based on these three dimensions, ELLC includes eight complexity levels: one basic task and five progressively more challenging tasks.

It should be noted that we benchmark extraction accuracy (symbolic transformation) and rearrangement accuracy (linguistic reasoning) as two tasks. For rearrangement accuracy, we report results by question (task) type: single, where the model is expected to generate one word and the sample is marked correct if the word matches any ground truth answer; and multiple, where the model must generate the complete list of all valid answers and is only considered correct if the list exactly matches the ground truth. Also, we emphasize a nuance in the multiple question setting: for samples with only one correct word, the model should infer this information without explicit guidance, making the task harder than when it is prompted to produce a single word. To enable finer-grained model benchmarking, we categorize performance into three groups: all samples, samples with a single correct answer, and samples with multiple correct answers.

\begin{table*}[t]
\resizebox{\textwidth}{!}{%
\begin{tabular}{cccccc}
\hline
\multicolumn{1}{c}{Number of letters} & Letter position         & Extraction accuracy    & Question type             & Sample type & Rearrangement accuracy \\ \hline
\multirow{12}{*}{4}                   & \multirow{6}{*}{Last}   & \multirow{6}{*}{0.976} & \multirow{3}{*}{Single}   & Overall     & 0.832            \\
&                         &                        &                           & Single      & 0.820                  \\
&                         &                        &                           & Multiple    & 0.982                  \\ \cline{4-6} 
&                         &                        & \multirow{3}{*}{Multiple} & Overall     & 0.638                  \\
&                         &                        &                           & Single      & 0.665                  \\
&                         &                        &                           & Multiple    & 0.309                  \\ \cline{2-6} 
& \multirow{6}{*}{Middle} & \multirow{6}{*}{0.615} & \multirow{3}{*}{Single}   & Overall     & 0.509                  \\
&                         &                        &                           & Single      & 0.500                  \\
&                         &                        &                           & Multiple    & 0.612                  \\ \cline{4-6} 
&                         &                        & \multirow{3}{*}{Multiple} & Overall     & 0.396                  \\
&                         &                        &                           & Single      & 0.415                  \\
&                         &                        &                           & Multiple    & 0.176                  \\ \hline
\multirow{12}{*}{6}                   & \multirow{6}{*}{Last}   & \multirow{6}{*}{0.863} & \multirow{3}{*}{Single}   & Overall     & 0.669                  \\
&                         &                        &                           & Single      & 0.662                  \\
&                         &                        &                           & Multiple    & 0.818                  \\ \cline{4-6} 
&                         &                        & \multirow{3}{*}{Multiple} & Overall     & 0.430                  \\
&                         &                        &                           & Single      & 0.439                  \\
&                         &                        &                           & Multiple    & 0.225                  \\ \cline{2-6} 
& \multirow{6}{*}{Middle} & \multirow{6}{*}{0.334} & \multirow{3}{*}{Single}   & Overall     & 0.286                  \\
&                         &                        &                           & Single      & 0.286                  \\
&                         &                        &                           & Multiple    & 0.304                  \\ \cline{4-6} 
&                         &                        & \multirow{3}{*}{Multiple} & Overall     & 0.185                  \\
&                         &                        &                           & Single      & 0.189                  \\
&                         &                        &                           & Multiple    & 0.082                 \\\hline
\end{tabular}%
}
\caption{Benchmark performance of GPT-4o on ELLC tasks with varying complexity. Structured output is achieved by format-restricting instruction.}
\label{ellc_benchmark_result}
\end{table*}

To establish an initial benchmark, Table \ref{ellc_benchmark_result} reports GPT-4o's accuracy on extraction and rearrangement across all ELLC tasks using the standard instruction and JSON format. First, GPT-4o performs less accurately on rearrangement than on extraction (the original LLC task), reflecting the additional linguistic reasoning introduced by ELLC. Second, when prompted to generate a candidate answer for questions with multiple correct answers, GPT-4o performs better, as it is more easier to produce at least one correct answer in these cases. Third, for samples with a single correct answer, GPT-4o performs more accurately when explicitly prompted to generate a single answer instead of when it is told there may be multiple correct answers — reflecting the confusion LLMs experience when there is uncertainty about the number of answers. Fourth, for samples with multiple correct answers, GPT-4o's performance drops significantly due to their greater complexity. Fifth, the middle-letter position is more challenging than the last-letter position, which may be influenced by overfitting stemming from LLC's popularity; in addition, the middle-letter condition is inherently more complex because it requires LLMs to handle subclasses of words with odd or even numbers of letters. Fifth, GPT-4o performance drops as word length increases from four letters to six letters, reflecting the growing complexity introduced by additional words. Overall, ELLC offers a challenging benchmark with varying complexity levels, where the state-of-the-art model achieves less than 10\% accuracy on the most difficult task. This makes it a valuable arena for evaluating LLMs' capabilities in both symbolic and linguistic reasoning.

\section{Concluding remarks}
The generation of LLMs has become more than a technological breakthrough; it has blossomed into a self-contained cosmos, rich with complexity and potential. As we pilot our vessels through this vast and ever-expanding cosmos, we often find ourselves orbiting familiar galaxies, composed of benchmarks, evaluation metrics, and empirical findings that suggest a similar scene. Yet what may seem like the entire universe is, in truth, only a dense constellation of closely aligned stars. To chart a truer and more comprehensive map of this universe, we turn to a foundational and enduring compass: causal inference, navigate the less traversed regions about structured output format, and document the nuance and inhomogeneity of LLMs cosmos. Through our experiments on the reasoning capabilities of LLMs, we reveal a multifaceted impact influenced by various factors. We emphasize that the proposed analytical pipeline can be readily extended to investigate additional inner parameters or external modules affecting LLMs' generation, supporting impact assessment and regulatory documentation in highly regulated industries.

\end{document}